\DeclareUnicodeCharacter{202F}{\,}
\documentclass[sigconf,screen]{acmart}

\usepackage{balance}

\AtBeginDocument{%
  \providecommand\BibTeX{{%
    \normalfont B\kern-0.5em{\scshape i\kern-0.25em b}\kern-0.8em\TeX}}}

\setcopyright{rightsretained}
\acmPrice{}
\acmDOI{10.1145/3468264.3468611}
\acmYear{2021}
\copyrightyear{2021}
\acmSubmissionID{fse21main-p691-p}
\acmISBN{978-1-4503-8562-6/21/08}
\acmConference[ESEC/FSE '21]{Proceedings of the 29th ACM Joint European Software Engineering Conference and Symposium on the Foundations of Software Engineering}{August 23--28, 2021}{Athens, Greece}
\acmBooktitle{Proceedings of the 29th ACM Joint European Software Engineering Conference and Symposium on the Foundations of Software Engineering (ESEC/FSE '21), August 23--28, 2021, Athens, Greece}

\begin{document}

\title{Empirical Study of Transformers for Source Code}

\author{Nadezhda Chirkova}
\affiliation{%
  \institution{HSE University}
  \city{Moscow}
  \country{Russia}
}
\email{nchirkova@hse.ru}

\author{Sergey Troshin}
\affiliation{%
  \institution{HSE University}
  \city{Moscow}
  \country{Russia}}
\email{stroshin@hse.ru}

\renewcommand{\shortauthors}{N. Chirkova and S. Troshin}

\begin{abstract}
  Initially developed for natural language processing (NLP), Transformers are now widely used for source code processing, due to the format similarity between source code and text. In contrast to natural language, source code is strictly structured, i.e., it follows the syntax of the programming language. Several recent works develop Transformer modifications for capturing syntactic information in source code. The drawback of these works is that they do not compare to each other and consider different tasks. In this work, we conduct a thorough empirical study of the capabilities of Transformers to utilize syntactic information in different tasks. We consider three tasks (code completion, function naming and bug fixing) and re-implement different syntax-capturing modifications in a unified framework. We show that Transformers are able to make meaningful predictions based purely on syntactic information and underline the best practices of taking the syntactic information into account for improving the performance of the model.
\end{abstract}





\maketitle

\section{Introduction}

Transformer~\cite{transformer} is currently a state-of-the-art architecture 
in a lot of source code processing tasks, including code completion~\cite{code-prediction-transformer}, code translation~\cite{code_translation, tree_encoding}, and bug fixing~\cite{Hellendoorn}. Particularly, Transformers were shown to outperform classic deep learning architectures, e.g., recurrent (RNNs), recursive and convolutional neural networks in the mentioned tasks. These architectures focus on \emph{local} connections between input elements, while Transformer processes all input elements in parallel and focuses on capturing \emph{global} dependencies in data, producing more meaningful code representations~\cite{Hellendoorn}. This parallelism also speeds up training and prediction.

Transformer is often applied to source code directly, treating code as  a sequence of language keywords, punctuation marks, and identifiers. In this case, a neural network mostly relies on identifiers, e.\,g.\,variable names, to make predictions~\cite{subroutines,summarisation}. High-quality variable names can be a rich source of information about the semantics of the code; however, this is only an indirect, secondary source of information. The primary source of information of what the code implements is its syntactic structure.

Transformer architecture relies on the self-attention mechanism that is not aware of the order or structure of input elements and treats the input as an unordered \emph{bag} of elements. To account for the particular structure of the input, additional mechanisms are usually used, e.g. positional encoding for processing sequential structure. In recent years, a line of research has developed mechanisms for utilizing tree structure of code in Transformer~\cite{tree_encoding, code-prediction-transformer, Hellendoorn}. However, the most effective way of utilizing syntactic information in Transformer is still unclear  for  three reasons. First,  the  mechanisms were developed concurrently, so they were not compared to each other by their authors. Moreover, different works test the proposed mechanisms on different code processing tasks, making it hard to align the empirical results reported in the papers. Secondly, the mentioned works used standard Transformer with positional encodings as a baseline, while modern practice uses more advanced modifications of Transformer, e.g., equipping it with relative attention~\cite{seq_rel_att}. As a result, it is unclear whether using sophisticated mechanisms for utilizing syntactic information is needed at all. Thirdly, most of the works focus on utilizing tree structure in Transformer and do not investigate the effect of processing other syntax components, e. g. the syntactic units of the programming language. 

\begin{figure*}[h!]
    \centering
         \includegraphics[width=\linewidth]{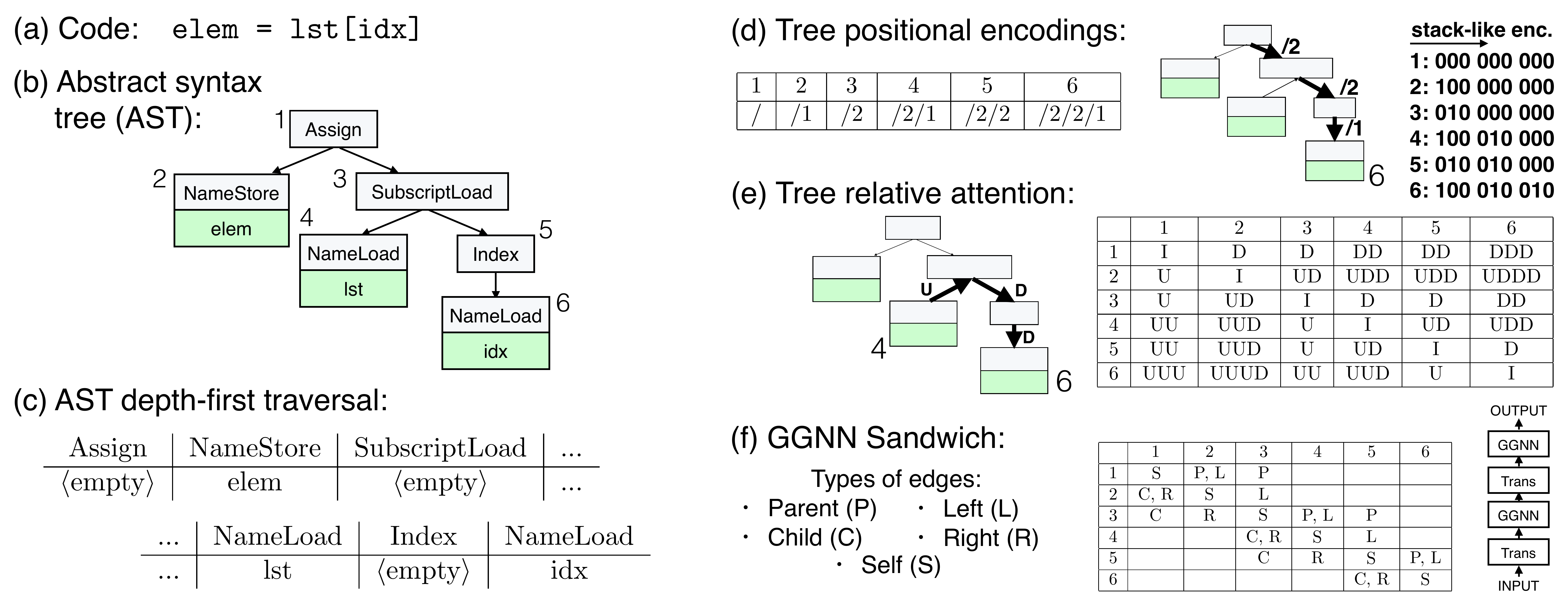} 
        \caption{Illustration of mechanisms for  processing AST structure in Transformer.}
        \label{fig:concepts}
\end{figure*}

In this work, we conduct an  empirical study of using Transformer for processing source code. Firstly, we would like to answer the question, what is the best way of utilizing syntactic information in Transformer, and to provide practical recommendations for the use of Transformers in software engineering tasks. Secondly, we aim at understanding whether Transformer is  generally suitable for capturing code syntax, to ground the future development of Transformers for code.
Our contributions are as follows:
\begin{itemize}
    \item We re-implement several approaches for capturing syntactic structure in Transformer and investigate their effectiveness in three code processing tasks on two datasets. We underline the importance of evaluating code processing models on several different tasks and believe that our work will help to establish standard benchmarks in neural code processing.
    \item We introduce an anonymized setting in which all user-defined identifiers are replaced with placeholders, and show that Transformer is capable of making meaningful predictions based purely on syntactic information, in all three tasks. We also show that using the proposed anonymization can improve the quality of the model, either a single Transformer or an ensemble of Transformers. 
    \item We conduct an ablation study of different syntax-capturing components in Transformer, underlining which ones are essential for achieving high quality and analysing the results obtained for the anonymized setting.
\end{itemize}
Our source code is available at \url{https://github.com/bayesgroup/code_transformers}.

The rest of the work is organized as follows. In Section~\ref{sec:review} we review  
the existing approaches for utilizing syntactic information in Transformer. In Section~\ref{sec:method} we describe our methodology for the empirical evaluation  of Transformer capabilities to utilize syntactic information. The following Sections~\ref{sec:exp1}--\ref{sec:exp3} describe our empirical findings. In Section~\ref{sec:litreview}, we give the review of the literature connected to our research. Finally, Section~\ref{sec:threats} discusses threats to validity and Section~\ref{sec:concl} concludes the work.

\section{Review of Transformers for Source code}
\label{sec:review}

\paragraph{Abstract syntax tree.}
A syntactic structure of code is usually  
represented in the form of 
an abstract syntax tree (AST).
Each node of the tree contains a type, which represents the syntactic unit of the programming language (e.g. "Assign", "Index", "NameLoad"), some nodes also contain a value (e.g. "idx", "elem"). Values store user-defined variable names, reserved language names, integers, strings etc. An example AST for a code snippet is shown in Figure~\ref{fig:concepts}(b).

\subsection{Transformer architecture and self-attention mechanism}
We describe Transformer architecture for a task of mapping input sequence $c_1, \ldots c_L$, $c_i \in \{1, \dots, M\}$ ($M$ is a vocabulary size) to a sequence of $d_{model}$-dimensional representations $y_1, \dots, y_L$ that can be used for making task-specific predictions in various tasks. Before being passed to the Transformer blocks, the input sequence is firstly mapped into a sequence of embeddings $x_1, \dots, x_L$, $x_i \in \mathbb{R}^{d_{model}}$.

A key ingredient of a Transformer block is a self-attention layer that maps input sequence $x_1, \ldots x_L$, $x_i \in \mathbb{R}^{d_{model}}$ to a sequence of the same length: $z_1, \ldots, z_L$, $z_i \in \mathbb{R}^{d_z}$. Self-attention first computes key, query, and value vectors from each input vector: $x^k_j = x_j W^{K}$, $x^q_j = x_j W^{Q}$ and $x^v_j = x_j W^{V}$. Each output $z_i$ is computed as a weighted combination of inputs:
\begin{equation}
\label{eq:att_matrix}
    z_i = \sum\limits_{j} \tilde \alpha_{ij} x^v_j, \quad \tilde \alpha_{ij} = \frac{\exp(a_{ij})}{\sum_j \exp(a_{ij})}, \quad
    a_{ij} =  \frac{x^q_i {x^k_j}^T}{\sqrt{d_z}}
\end{equation}
Attention weights $\tilde \alpha_{ij} \geqslant 0$, $\sum_{j=1}^L \alpha_{ij}=1$ are computed based on query-key similarities.
Several attention layers (heads) are applied in parallel with different projection matrices $W^V_h$, $W^Q_h$, $W^K_h$, $h=1, \dots, H$. The outputs are concatenated and projected to obtain $\hat{x}_i = [z_{i}^1, \ldots, z_{i}^H]W^O$, $W^O \in \mathbb{R}^{H d_z \times d_{model}}$. A Transformer block includes the described multi-head attention, a residual connection, a layer normalization, and a position-wise fully-connected layer.
The overall Transformer architecture is composed by the consequent stacking of the described blocks. When applying Transformer to generation tasks, future elements ($i>j$) are masked in self-attention (Transformer \emph{decoder}). Without this masking, the stack of the layers is called Transformer \emph{encoder}. In the sequence-to-sequence task, when both encoder and decoder are used, the attention from decoder to encoder is also incorporated into the model. 

\subsection{Passing ASTs to Transformers}
For our study, we select two commonly used NLP approaches for utilizing sequential structure and three approaches developed specifically for utilizing source code structure in Transformer.

\paragraph{Sequential positional encodings and embeddings.}
Transformers were initially developed for NLP and therefore were augmented with sequence-capturing mechanisms to account for sequential input structure. As a result, the simplest way of applying Transformers to AST is to traverse AST in some order, e.g., in depth-first order (see Figure~\ref{fig:concepts}(c)), and use standard sequence-capturing mechanisms.

To account for the sequential nature of the input, standard Transformer is augmented with positional encodings or positional embeddings. Namely, the input embeddings $x_i \in R^{d_{model}}$ are summed up with positional representations $p_i \in R^{d_{model}}$: $\hat{x}_i = x_i + p_i$. For example, positional embeddings imply learning the embedding vector of each position $i \in 1 \ldots L$:
$ p_i = e_i, \,\,e_i \in R^{d_{model}}$. Positional encoding implies computing $p_i$ based on sine and cosine functions.
We include positional embeddings in our comparisons and add the prefix ``sequential'' to the title of this mechanism.  This approach was used as a baseline in several of works~\cite{tree_encoding,Hellendoorn}.

\paragraph{Sequential relative attention.}
\citet{seq_rel_att} proposed relative attention for capturing the order of the input elements. 
They augment self-attention with relative embeddings:
    \begin{equation}
        z_i = \sum\limits_{j} \tilde \alpha_{ij} (x^v_j + e_{i-j}^v),\,\,\tilde \alpha_{ij} = \frac{\exp(a_{ij})}{\sum_j \exp(a_{ij})},
        \,\,
        a_{ij} =  \frac{x^q_i(x^k_j+ e_{i-j}^k)^T}{\sqrt{d_z}},
    \end{equation} where $e_{i-j}^v, e_{i-j}^k \in \mathbb{R}^{d_z}$ are learned embeddings for each relative position $i - j$, e.\,g.\,one token is located two tokens to the left from another token.
This mechanism, that we call \emph{sequential relative attention}, was shown to substantially outperform sequential positional embeddings and encodings in sequence-based text processing tasks.
\citet{summarisation} reach the same conclusion evaluating sequential relative attention in a task of code summarization, i.e., generating natural language summaries for code snippets.

\paragraph{Tree positional encodings.}
Inspired by previously discussed works, several  authors developed mechanisms for processing trees. 
\citet{tree_encoding} develop positional encodings for tree-structured data, assuming that the maximum number $n_w$ of node children  and  the maximum depth $n_d$ of the tree are relatively small. Example encodings are given in Figure~\ref{fig:concepts}(d). The \emph{position} of each node in a tree is defined by its path from the root, and each child number in the path is encoded using $n_w$-sized one-hot vector. The overall representation of a node is obtained by concatenating these one-hot vectors in reverse order and padding short paths with zeros from the right. The authors also introduce the learnable parameters of the encoding, their number equals $d_{\mathrm{model}}/(n_w\cdot n_d)$. Paths longer than $n_d$ are clipped (the root node is clipped first). The authors binarize ASTs to achieve $n_w=2$. To avoid this binarization, we replace all child numbers greater than $n_w$ with $n_w$, and select the best hyperparameters $n_w$ and $n_d$ using grid search, see details in section~\ref{sec:exp_setup}.

\citet{tree_encoding} tested the approach on the task of code translation (code-to-code) and semantic parsing (text-to-code). The Transformer with tree positional encodings outperformed standard Transformer with sequential positional encodings and TreeLSTM \cite{treelstm}.
    
\paragraph{Tree relative attention.}    
An extension of sequential relative attention for trees was proposed by
\citet{code-prediction-transformer}. In a sequence, the distance between two input positions is defined as the number of positions between them. Similarly, in a tree, the distance between two nodes can be defined as the shortest path between nodes, consisting of $n_{U} \geqslant 0$ steps \emph{up} and $n_{D} \geqslant 0$ steps \emph{down}, see example in Figure~\ref{fig:concepts}(e). Now, similarly to sequential relative attention, we can learn embeddings for the described distances and plug them into self-attention. Learning multidimensional embeddings for the tree input requires much more memory than for sequential input, since distances in the tree are object-specific, while distances in the sequence are the same for all objects in a mini-batch. As a result, the authors use scalar embedding $r_{ij} \in \mathbb{R}$ for the distance between nodes $i$ and $j$ and plug it into the attention mechanism as follows (other formulas stay the same):
\begin{equation}
        \tilde \alpha_{ij} = \frac{\exp(a_{ij} \cdot r_{ij})}{\sum_j \exp(a_{ij} \cdot r_{ij})}.
    \end{equation} 
Our preliminary experiments suggested that using summation $\alpha_{ij}+r_{ij}$ instead of multiplication leads to a higher final score.
The authors tested the approach
     on the task of code completion, i.e., predicting the next token, and showed that using modified attention improves quality when applying Transformer to AST traversal and to code as text.

\paragraph{GGNN Sandwich.}
Due to the graph nature of AST, source codes are often processed using graph gated neural networks (GGNN)~\cite{allamanis_iclr2018}. To add more inductive bias, AST is augmented with edges of several additional types, e.g., reflecting data- and control-flow in the program. Such a model captures \emph{local} dependencies in data well but lacks a \emph{global} view of the input program, that is the Transformer's forte. Inspired by this reasoning,
\citet{Hellendoorn} propose alternating  Transformer and GGNN layers as illustrated in Figure~\ref{fig:concepts}(f), to combine the strengths of both models. GGNN layer relies on passing messages through edges for a fixed number of iterations (number of passes). The model is called GGNN Sandwich by the authors, and the details can be found in~\cite{Hellendoorn}.
GGNN Sandwich was shown to be effective in the variable misuse detection task, i.e., predicting the location of a bug and the location used to fix the bug (copy variable). GGNN Sandwich outperformed standard Transformer with sequential positional encodings. 

Our work focuses of processing syntactic information in Transformer, thus we do not use data- and control-flow edges. Data- or control-flow edges are hard to incorporate in other mechanisms except GGNN Sandwich. In our GGNN Sandwich, we use AST edges, edges connecting the neighbouring nodes in the AST depth-first traversal, and edges connecting nodes to themselves, see illustration in fig.~\ref{fig:concepts}(f). 

\citet{Hellendoorn} also propose a model called GREAT that is inspired by relative attention and incorporates 1-dimensional edge embeddings into the attention mechanism. This model is conceptually similar to the tree relative attention, thus we do not include GREAT in our comparison.

\section{Limitations of existing approaches and methodology of the work}
\label{sec:method}
As shown in Section~\ref{sec:review} several approaches for processing ASTs in Transformers have been proposed. However, it is still unclear which approaches perform better than others and what mechanisms to use in practice. 
First, all the works discussed in Section~\ref{sec:review} conduct experiments with different tasks making it hard to align the results.
Moreover, almost all the listed works compare their approaches with the vanilla Transformer, i.e., Transformer with sequential positional encodings or embeddings, while modern practices  use advanced mechanisms, like sequential relative attention, by default. Even works that propose tree-processing approaches inspired by sequential relative attention do not include this mechanism as a baseline. That is, it is unclear whether using advanced tree-processing mechanisms is beneficial at all.
Secondly, the existing approaches focus on capturing tree \emph{structure} and do not investigate the influence of other components of AST, i.e., types and values. 

In this work, we conduct a thorough empirical study on utilizing AST in Transformers. We consider three code processing tasks: variable misuse (VM) detection, function naming (FN), and code completion (CC), and two source code datasets: Python150k~\cite{python_dataset} and JavaScript150k~\cite{javascript150k}. We selected tasks that are often used as benchmarks in the literature and on which the compared approaches were tested by their authors. Our selection also covers various Transformer configurations, i.e., encoder only (VM), decoder only (CC) and encoder-decoder (FN). We selected the Python150k dataset because it is often used in the literature, and JavaScript150k because it is distributed by the same authors and has the same format.

We re-implement all mechanisms described in Section~\ref{sec:review} in a unified framework and investigate the most effective approach for processing ASTs in Transformer in different tasks. We answer the following research questions:
\begin{itemize}
     \item What is the most effective approach for utilizing AST \emph{structure} in Transformer?
    \item Is Transformer generally capable of utilizing syntactic information represented via AST?
    \item What components of AST (structure, node types and values) does Transformer use in different tasks?
\end{itemize}
There is no common practice of preprocessing ASTs, particularly, processing values. Each node in AST is associated with a type, but not all nodes have associated values.  \citet{code-prediction-transformer} and \citet{tree_encoding} attach values as separate child nodes so that each node stores only one item (type or value), while \citet{Hellendoorn} propose omitting types.
The former approach increases input length and thus makes code processing significantly slower, while the latter approach loses type information. We choose an in-between strategy inspired by the approach of~\citet{pointer} used for RNNs: we associate the \verb|<empty>| value with nodes that do note have values, so that each node $i$ in AST has both type $t_i$ and value $v_i$, see Figure~\ref{fig:concepts}(c).  This setup preserves the initial AST structure and allows us to easily ablate types or values, leaving the other item in each node present. 

Some works show that splitting values based on \verb|snake_case| or \verb|CamelCase|, or using splitting techniques such as byte-pair encoding may improve the quality~\cite{bigcode,code2seq}. We do not use splitting into subtokens for two reasons. Firstly, splitting makes sequences much longer, resulting in a substantial slow down of training procedure because of quadratic Transformer complexity w.r.t. the input length.
Secondly, splitting breaks the one-to-one correspondence between AST nodes and values, i.e., several values belong to one AST node. There are different ways of adapting AST-based Transformers to the described problem: one option is to average embeddings over subtokens~\cite{Hellendoorn}, another option is to assign a chain of subtokens as a child of a node and then directly apply tree-processing mechanisms. A third option is to modify tree-processing mechanisms, e.g., duplicate paths for all subtokens in tree positional encoding or duplicate tree relations for all pairs of subtokens of two tokens in tree relative attention. As a result, the question of how splitting into subtokens affects syntax-capturing mechanisms requires a separate study which we leave for the future work.

An important part of our methodology is conducting experiments in two settings, namely \emph{anonymized} and \emph{full-data}. The full-data setting corresponds to the conventional training of Transformer on ASTs parsed from code. In this case, Transformer has two sources of information about input code snippets: syntactic information and user-defined identifiers (stored in node values). Identifiers usually give much \emph{additional} information about the semantics of the code, however, their presence is not necessary for correct code execution: renaming all user-defined identifiers with placeholders \verb|var1|, \verb|var2|, \verb|var3| etc. will lead to the same result of code execution and will not change the semantics of the algorithm the code implements. Here we mean that all occurrences of an identifier are replaced with the same placeholder, thus, important information about identifier repetition is saved. We call this renaming identifiers with placeholders as \emph{anonymization}.
In the anonymized setting, the input code is represented purely with syntax structure and the only way Transformer can make meaningful predictions is to capture information from AST. In this way, using the anonymized setting allows a better understanding of the capabilities of Transformer to utilize syntactic information. More details on the anonymization procedure are given in Appendix~\ref{app:hypers}. 

Another important part of our methodology is a thoughtful  splitting of the dataset into training and testing sets, which includes splitting by repository and removing code duplicates. \citet{code2seq, subroutines} notice that code files inside one repository usually share variable names and code patterns, thus splitting files from one repository between training and testing sets simplifies predictions for the testing set and leads to a data leak. To avoid this, one should put all files from one repository into one set, either training or testing (this strategy is called splitting by repository). Even using  this strategy, duplicate code can still occur in the testing set, since developers often copy code from other projects or fork other repositories. 
\citet{deduplication} underline that in commonly used datasets up to 20\% of testing objects can be repeated in the training set, biasing evaluation results. As a result, the deduplication step is needed after data splitting.

\section{Experimental setup}
\label{sec:exp_setup}
\paragraph{Data.} In all tasks, we use the Python150k (PY) dataset ~\cite{python150k} (redistributable version) and JavaScript150k (JS) dataset~\cite{javascript150k} downloaded from the official repository at \url{https://eth-sri.github.io}. Both datasets consist of code files downloaded from Github and are commonly used to evaluate code processing models. 

Most research use the train-test split provided by the authors of the dataset, however, this split does not follow best practices described in Section~\ref{sec:method} and produce biased results~\cite{deduplication}, so we release a new split of the dataset. 
We remove duplicate files from both datasets using the list of duplicates provided by~\citet{deduplication}. We also filter out absolutely identical code files, and when selecting functions from code files, we additionally filter out absolutely identical functions. We split data into training~/~validation~/~testing sets in proportion 60\%~/~6.7\%~/~33.3\% based on Github usernames (each repository is assigned to one username).

Preprocessing details for each task are given below. We release our data split and our source code including scripts for downloading data, deterministic code for data preprocessing, models, training etc. We largely rely on the implementations of other research, and compare the quality of our baseline models to the results reported in other papers, when possible; see details in Section ~\ref{sec:valid}.

\paragraph{Variable misuse task (VM)} 
For the variable misuse task, we use the setup and evaluation strategy  of~\citet{Hellendoorn}. 
Given the code of a function, the task is to identify two positions (using two pointers): one in which position a wrong variable is used, and one in which position a correct variable can be copied from (any such position is accepted). If a snippet is non-buggy, the first pointer should select a special no-bug position. We obtain two pointers, by applying two position-wise fully-connected layers, and softmax over positions on top of Transformer outputs. 
For example, the first pointer selects position as $\text{argmax}_{1 \leqslant i \leqslant  L}\text{softmax}([u^T y_1, \dots, u^T y_L, b])$, $y_i \in \mathbb{R}^{d_{\mathrm{model}}},\, u \in \mathbb{R}^{d_{\mathrm{model}}}$, $b \in \mathbb{R}$ ($b$ is a learnable scalar corresponding to the no-bug position), $[\dots]$ denotes the concatenation of the elements into a vector of scalars. The second pointer is computed in a similar way but without $b$. The model is trained using  the cross-entropy loss.

To process the original dataset for the variable misuse task, we select all top-level functions, including functions inside classes, from all (filtered) 150K files, and filter out functions: longer than 250 nodes (to avoid very long functions); and functions with less than three  positions containing user-defined variables or less than three distinct user-defined variables (to avoid trivial bug fixes). We select a function with a root node type \verb|FunctionDef| for PY, and \verb|FunctionDeclaration| or \verb|Function| \verb|Expression| for JS. The resulting training~/~validation~/~testing set consists of 417K~/~48K~/~231K functions for PY and 202K~/~29K~/~108K for JS. One function may occur in the dataset up to 6 times, 3 times with a synthetically generated bug and 3 times without a bug. Following~\cite{Hellendoorn}, we use this strategy to avoid biasing towards long functions with a lot of different variables. The buggy examples are generated synthetically by choosing random bug and fix positions from positions containing user-defined variables. 

We use the joint localization and repair accuracy metric of~\cite{Hellendoorn} to assess the quality of the model. This metric estimates the portion of buggy samples for which the model correctly localizes and repairs the bug.
We also measured localization accuracy and repair accuracy independently and found that all three metrics correlate well with each other.

\paragraph{Function naming task (FN)} 
In this task, given the code of a function, the task is to predict the name of the function. To solve this task, we use the classic sequence-to-sequence Transformer architecture that outputs the function name word by word. A particular implementation is  borrowed from~\cite{summarisation} (the paper used another dataset).
Firstly,  we pass function code to the Transformer encoder to obtain code representations $y_1, \dots, y_L$. Then, the Transformer decoder generates the method name word by word, and during each word generation, decoder attends to $y_1, \dots, y_L$ (using encoder-decoder attention) and to previously generated tokens (using masked decoder attention). To account for the sequential order of the function name, we use sequential positional embeddings in the Transformer decoder. In the encoder, we consider different structure-capturing mechanisms. We use greedy decoding. We train the whole encoder-decoder model end-to-end, optimizing the cross-entropy loss.

To obtain the processed dataset for the function name task, we select all top-level functions, including functions inside classes, from all (filtered) 150K files, and filter out functions longer than 250 AST nodes (to avoid very long functions), functions for which the name could not be extracted (a lot of \verb|FunctionExpression|s in JS are anonymous), and functions with names consisting of only underscore characters and names containing rare words (less than 5~/~3 occurrences in the training set for PY~/~JS). To extract functions, we use the same root node types as in the VM task.  The resulting dataset consists of 523K~/~56K~/~264K training/validation/testing functions for PY and 186K~/~23K~/~93K for JS. We replace function name in the AST with a special \verb|<fun_name>| token. To extract target function names, we remove extra underscores and split each function name based on \emph{CamelCase} or \emph{snake\_case}, e.g., name \verb|_get_feature_names| becomes [\verb|get|, \verb|feature|, \verb|names|]. A mean$\pm$std length of function name is 2.42$\pm$1.46 words for PY and 2.22$\pm$1.23 for JS.

We assess the quality of the generated function names using the F1-metric. If $gtn$ is a set of words in ground-truth function name and $pn$ is a set of words in predicted function name, the F1-metric is computed as $2PR/(P+R) \in [0, 1]$, where $P=|gtn\cap pn|/|pn|$, $R = |gtn \cap pn| / |gtn|$, $|\cdot|$ denotes the number of elements. F1 is  averaged over functions. We choose the F1 metric following~\citet{code2seq} who solved a similar task with another dataset and model.

\paragraph{Code completion task (CC)}
For the task of code completion, we use the setup, metrics and Transformer implementation of~\citet{code-prediction-transformer}. The task is to predict the next node $(t_i, v_i)$ in the depth-first traversal of AST $[(t_1, v_1), \dots, (t_{i-1}, v_{i-1})]$. 
We predict type $t_i$ and value $v_i$ using two fully connected layers with softmax on top of the prefix representation $y_i$: $P(t_i) = \mathrm{softmax}(W^t y_i)$, $W^t \in \mathbb{R}^{\mathrm{\# types} \times d_{\mathrm{model}}}$, $P(v_i) = \mathrm{softmax}(W^v y_i)$, $W^v \in \mathbb{R}^{\mathrm{\# values} \times \mathrm{d_{\mathrm{model}}}}$.

To obtain the dataset for the code completion task, we use full ASTs from (filtered) 150k files, removing sequences with length less than $2$. If the number of AST nodes is larger than $n_{ctx}=500$, we split AST into overlapping chunks of length $n_{ctx}$ with a shift $\frac{1}{2}n_{ctx}$. The overlap provides a context for the model. For example, if the length of AST is $800$, we select the following samples: $AST[:\!500), AST[250\!:\!750), AST[300\!:\!800]$. We do not calculate loss or metrics over the intersection twice. For the previous example, the quality of predictions is measured only on $AST[:\!500), AST[500\!:\!750), AST[750\!:\!800]$. The overlapping splitting procedure is borrowed from~\cite{code-prediction-transformer} and is needed since processing extremely long sequences is too slow in Transformer because of its quadratic complexity w.r.t. input length. The resulting dataset consists of 186K~/~20K~/~100K training~/~validation~/~testing chunks for PY and 270K~/~32K~/~220K for JS.

We mostly focus on value prediction, since it is the more complex task, and present results for type prediction where they significantly differ from other tasks.
We optimize the sum of cross-entropy losses for types and values. 

We use mean reciprocal rank (MRR) to measure the model quality since it reflects the practical application of code completion: 
$ \mathrm{MRR} = \frac{1}{N} \sum_{i=1}^N 1/rank_i$, where $rank_i$ is a position of the $i$-th true token in the model ranking, $N$ is the total number of target tokens in a dataset, excluding \verb|<padding>| and \verb|<empty>| tokens. As in ~\cite{code-prediction-transformer}, we assign zero score if the true token is out of top $10$ predicted tokens.

\paragraph{Hyperparameters.}
We list general hyperparameters for the variable misuse / function naming / code completion tasks using slashes. Our Transformer models have 6 layers, 6~/~6~/~8 heads, $d_{model}=512$. We limit vocabulary sizes for values up to 50K / 50K / 100K tokens and preserve all types. As discussed in Section~\ref{sec:method}, we do not split values into subtokens. We train all Transformers using Adam with a starting learning rate of 0.00001~/~0.0001~/~0.0001 and a batch size of 32 for 25~/~15~/~20 epochs for PY and 40~/~25~/~20 epochs for JS (the number of functions in the JS dataset is smaller than in PY dataset, thus more epochs are needed). In the code completion task, we use the cosine learning rate schedule \cite{cosine} with 2000 warm-up steps and a zero minimal learning rate, and a gradient clipping of 0.2. In the variable misuse task and in the function naming task for JS, we use a constant learning rate. In the function naming task for PY, we decay the learning rate by 0.9 after each epoch. In the function naming task, we also use a gradient clipping of 5. We use residual, embedding and attention dropout with $p=0.2~/~0.2~/~0.1$.  All models were trained three times, to estimate the standard deviation of the quality (except hyperparameter tuning). In all experiments we report the quality on the test set, except hyperparameter tuning where we report the quality on the validation set. We train all models on one GPU (NVIDIA Tesla P40 or V100).

\begin{table}[t!]
	\caption{Selected hyperparameters for different structure-capturing mechanisms, tasks and datasets. The details on selecting hyperparameters are given in Appendix~\ref{app:hypers}.}\label{tab:hyper_choices}
	\begin{tabular}{c|c|cccc}
		Model (hypers.) & Lang. & VM & FN  & CC \\  \midrule
	    Seq. rel. attn. & PY & 8 & 250  & 32 \\
	    (max. dist.) & JS & 8 & 250 & 32 \\ \hline
	    Tree pos. enc. & PY & 8, 16 & 16, 8 & 16, 8 \\
	    (max width, depth) & JS & 4, 8  & 2, 64 & 16, 32 \\ \hline
	    Tree rel. attn. & PY & 100 & 600 & 1000 \\
	    (rel. vocab. size) & JS & 600  & 100 & 1000 \\ \hline
	    GGNN Sandwich & PY & 12, 3, N & 6, 2, Y & N/A \\
	    (num. layers, & JS & 12, 3, N & 6, 2, Y & N/A \\
	    num. edge types, & & & \\
	    is GGNN first?) & & & \\ \hline
	\end{tabular}
\end{table}

\begin{figure*}[t!]
\begin{tabular}{cccc}
         \hspace{0.2cm} Python: \hspace{0.55cm} Variable Misuse &  Function Naming &  Code Completion (values) & Code Completion (types)\\
          \includegraphics[height=0.15\linewidth]{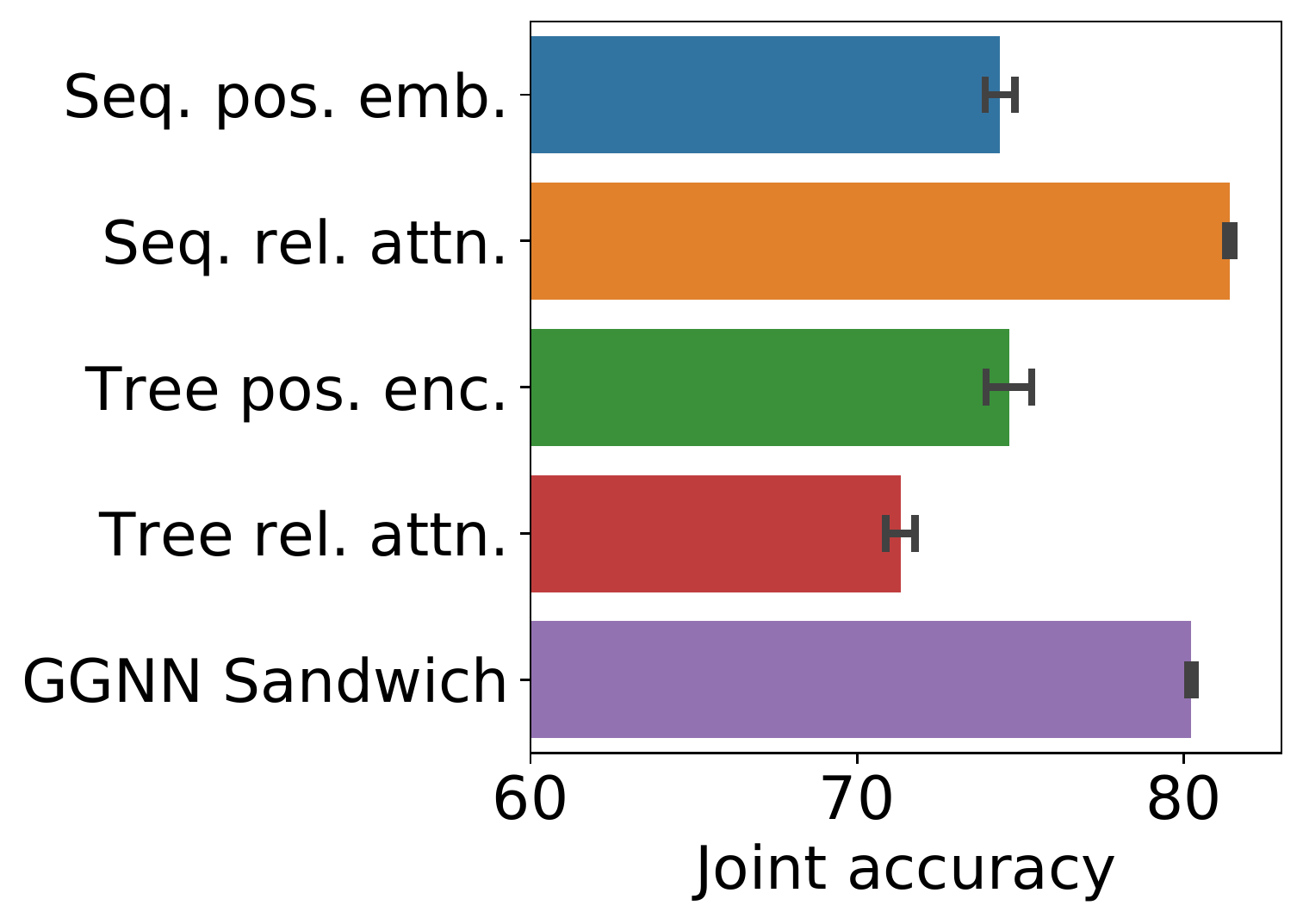} &
         \includegraphics[height=0.15\linewidth]{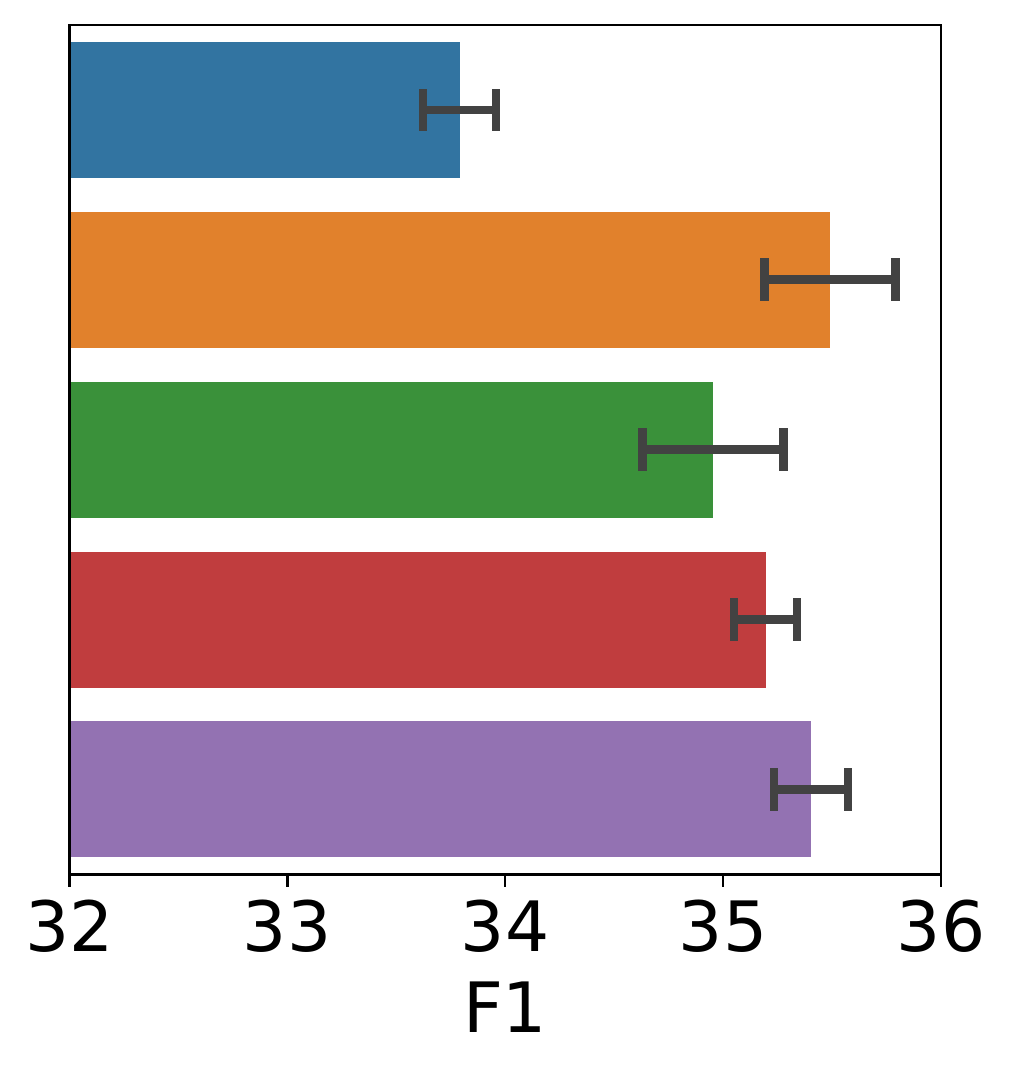} &
\includegraphics[height=0.15\linewidth]{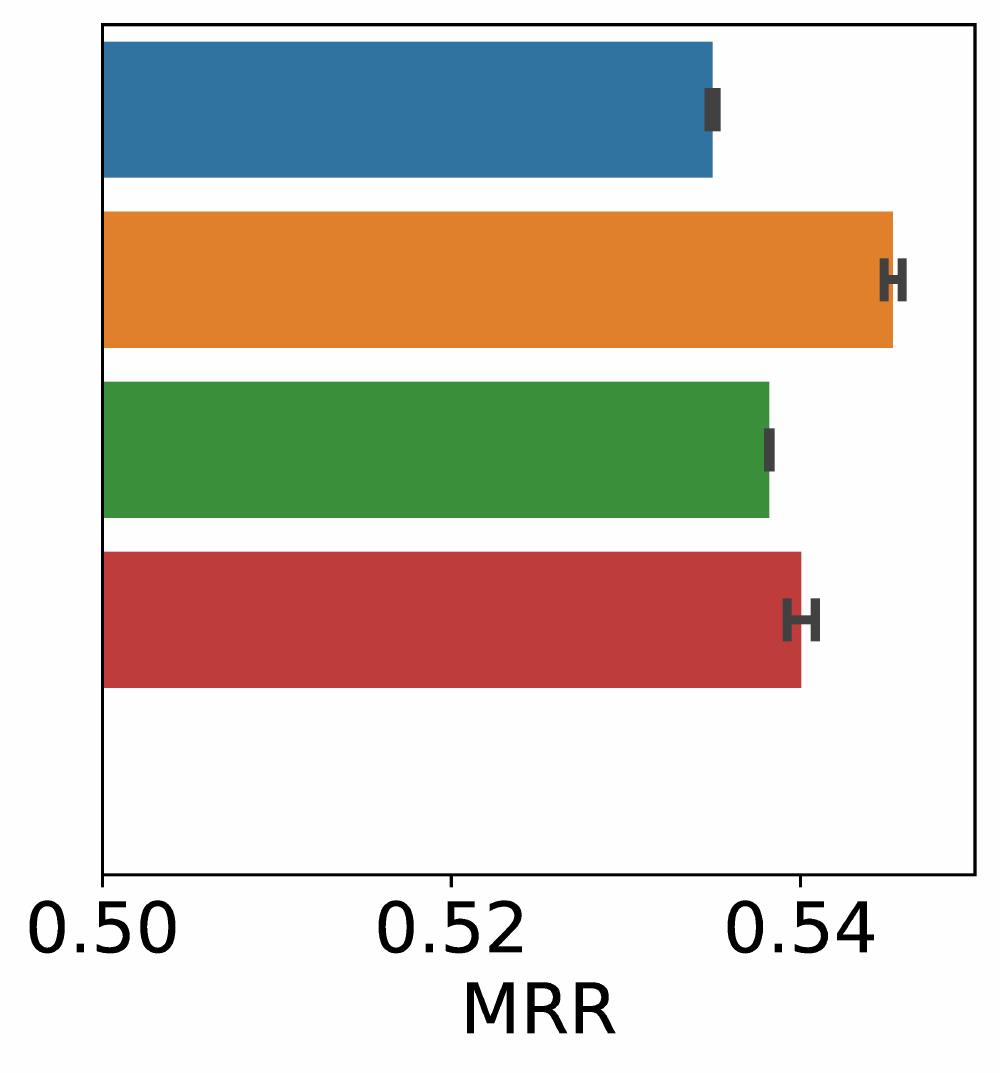}&
\includegraphics[height=0.15\linewidth]{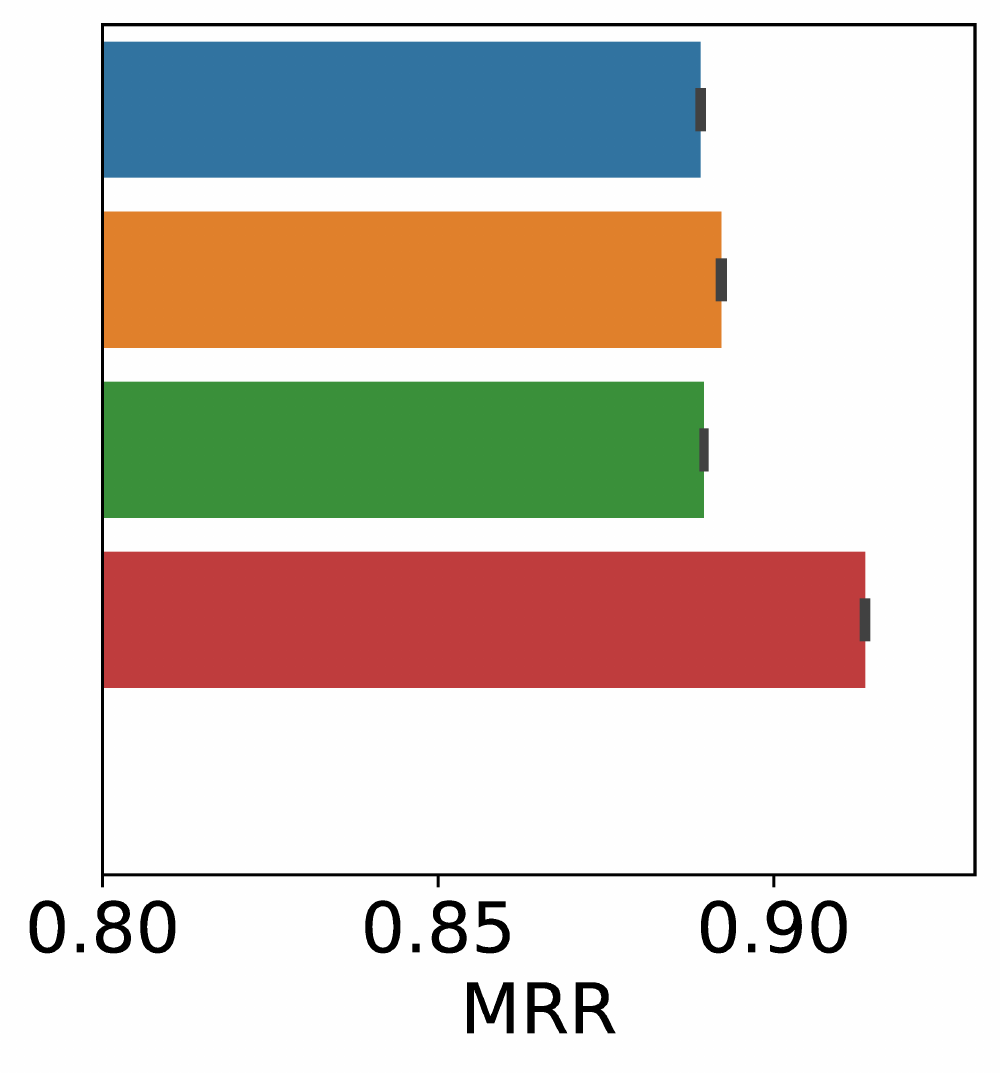}
\\
         JavaScript: \hspace{0.4cm} Variable Misuse \hspace{0.1cm} &  Function Naming  &  Code Completion (values)   &  Code Completion (types)  \\
         \includegraphics[height=0.15\linewidth]{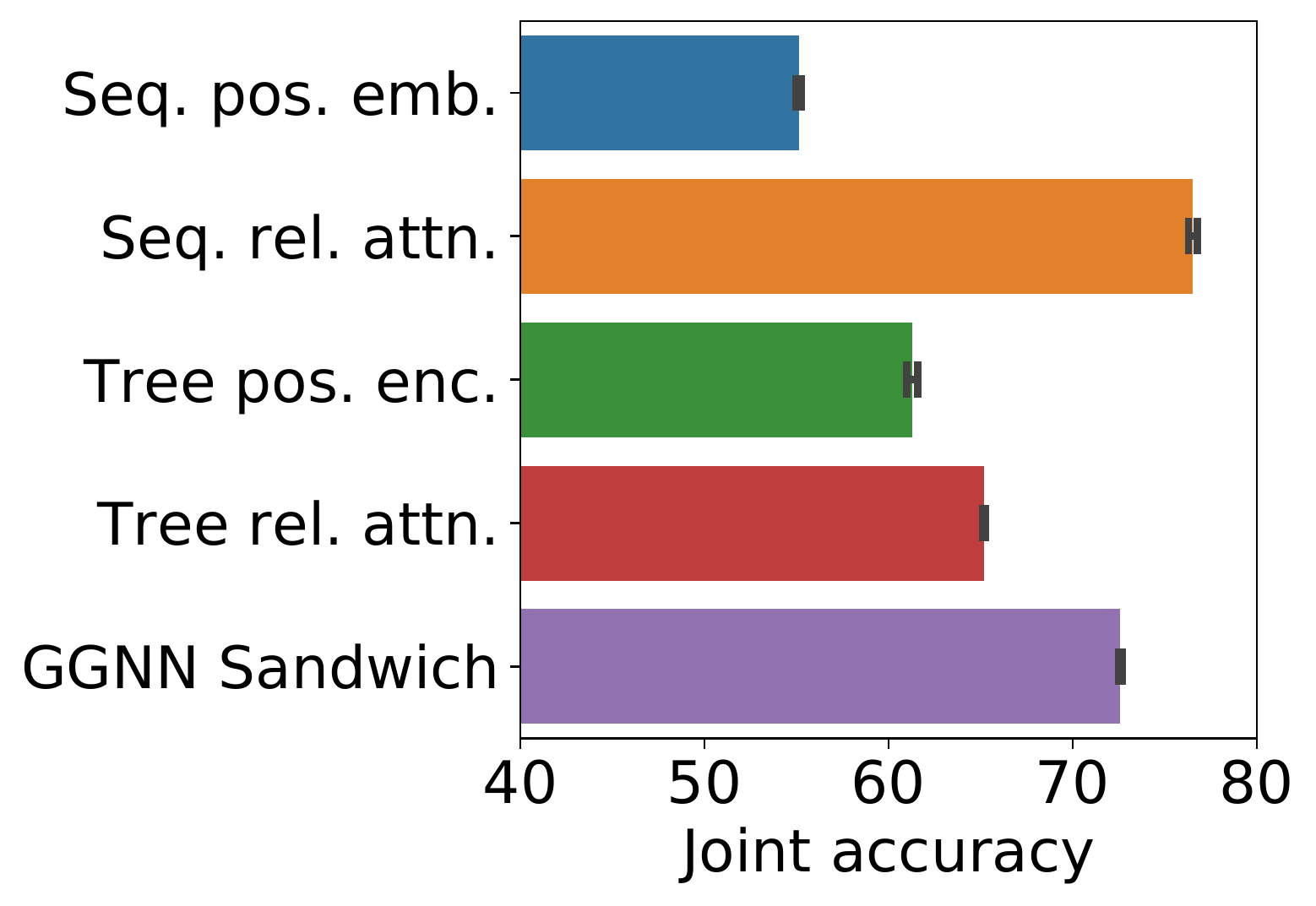} &
         \includegraphics[height=0.15\linewidth]{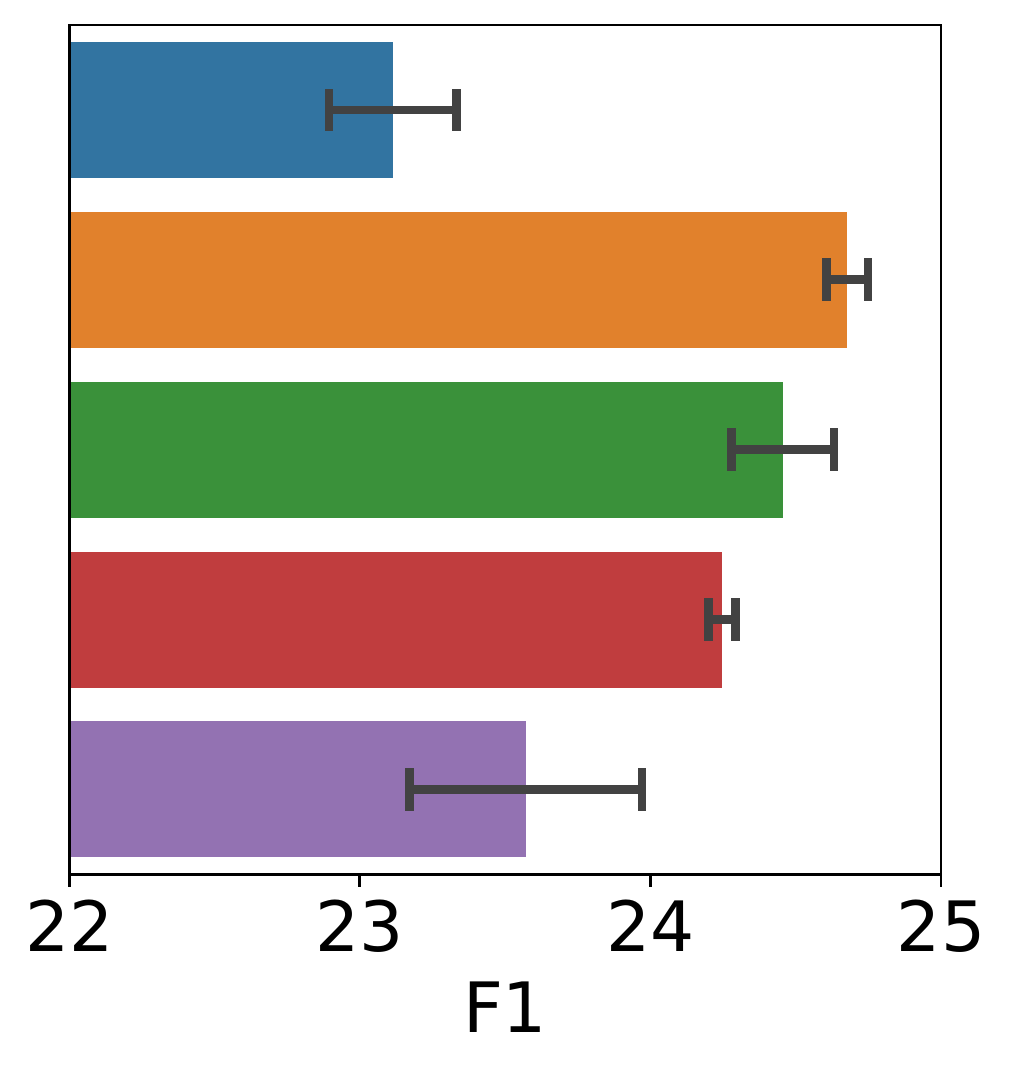} &
         \includegraphics[height=0.15\linewidth]{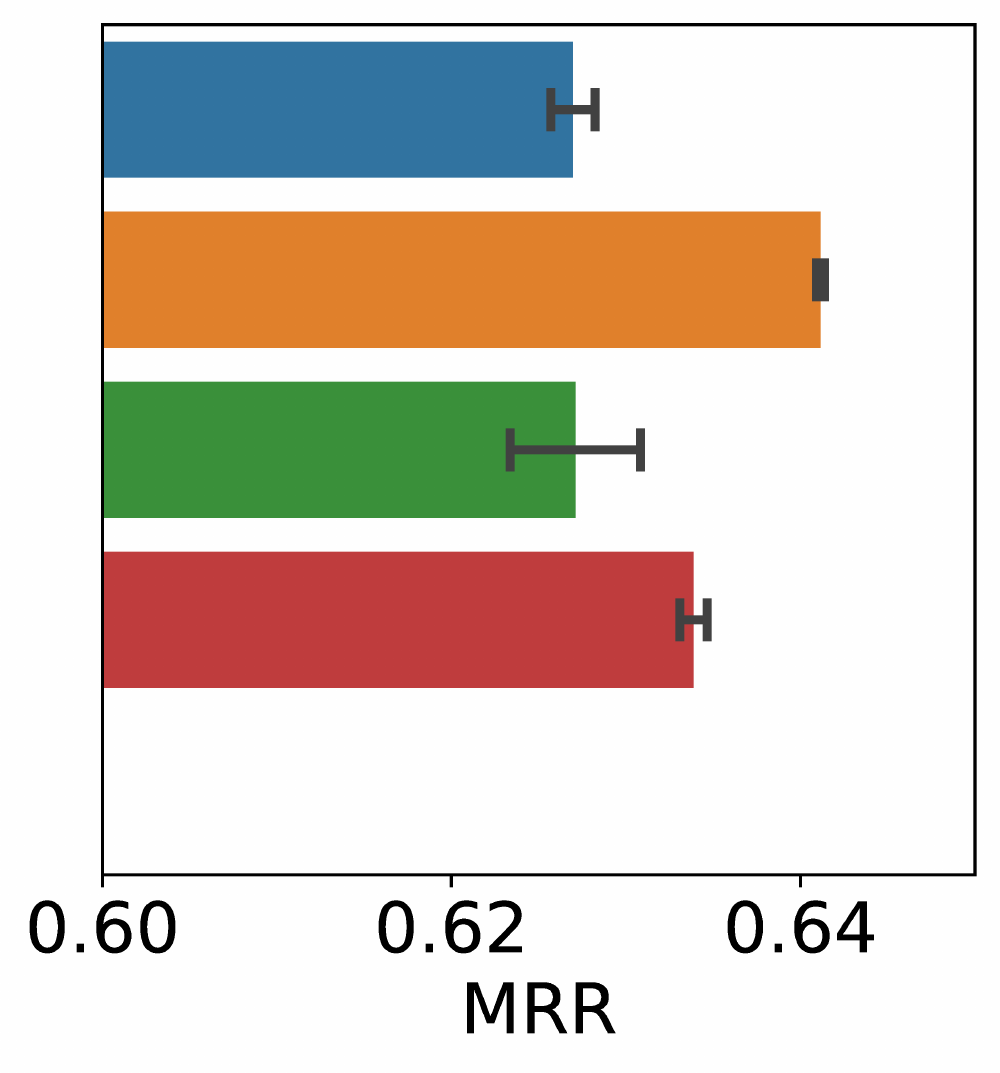} &
         \includegraphics[height=0.15\linewidth]{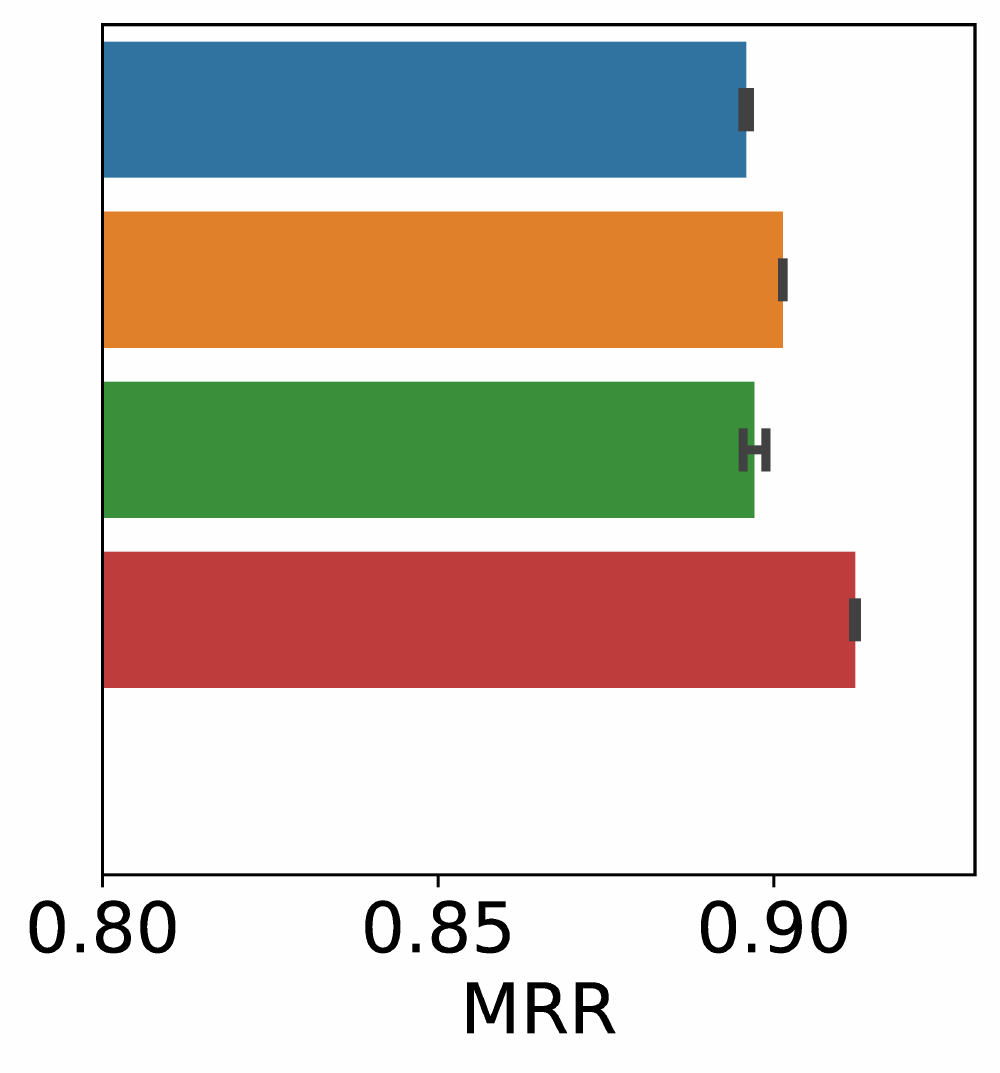}
         
\end{tabular}
\caption{A comparison of different mechanisms for processing AST structure in Transformer, in the full-data setting. The numeric data for barplots is given in Appendix~\ref{app:numdata}.}
\label{fig:exp2a_full}
\end{figure*}

The hyperparameters for different structure-capturing mechanisms were tuned using grid search, based on the quality on the the validation set, for each dataset--task combination individually. For sequential relative attention, we tune the maximum relative distance between the elements of the sequence. For tree positional encoding, we tune the maximum path width and the maximum path depth. For tree relative attention, we tune the size of the relation vocabulary. For the GGNN Sandwich model, we consider 6-layer and 12-layer configurations of alternating Transformer (T) and GGNN (G) layers, we also consider placing both types of layers first i.e., [T, G, T, G, T, G] or [G, T, G, T, G, T] (and similarly for 12 layers). GGNN layers include 4 message passes. We also consider omitting edges of types \verb|Left| and \verb|Right|. Sequential positional embeddings do not have hyperparameters. The number of parameters in all AST-based modifications of Transformer are approximately the same, except GGNN Sandwiches: 12-layer Sandwich incorporates slightly more parameters than vanilla Transformer, while 6-layer incorporates slightly fewer parameters. The details and the Tables on hyperparameter search are given in Appendix~\ref{app:hypers}, the resulting hyperparameters are listed in Table~\ref{tab:hyper_choices}.

\section{Comparison of approaches for utilizing syntactic structure in Transformer}
\label{sec:exp2a}
We begin with investigating which of the mechanisms for utilizing AST structure in Transformer is the most effective one.
We obtain the trees storing a (type, value) pair in each node using the approach described in Section~\ref{sec:method} and pass these trees to Transformer, equipped with one of the mechanisms described in Section~\ref{sec:review}.
GGNN Sandwich is not applicable to code completion because message-passing involves all nodes and prohibits using masking in the decoder.

The results are presented in Figure~\ref{fig:exp2a_full}. In function naming, most structure-capturing mechanisms perform similarly. In Section~\ref{sec:exp2}, we show that in this task quality is not affected much even if we completely ablate structure information, i.e., do not use any structure-capturing mechanism and treat input as a \emph{set} of (type, value) pairs. That is, Transformer hardly utilizes syntactic structure when predicting function names.
However, in other tasks, this is not the case and there is more variability in different mechanisms performance. 

Utilizing structure information in the input embeddings is not effective: sequential positional embeddings and tree positional encodings do not achieve highest score in any task, except function naming where tree positional encodings perform on par with other mechanisms. 

Utilizing structure in the self-attention mechanism is much more effective: in all tasks, at least one of sequential relative attention and tree relative attention is the best performing model. Sequential relative attention achieves the highest score in variable misuse and value prediction tasks, while tree relative attention outperforms others by a high margin in type prediction task (this model was developed for code completion task). The last result is interpretable since tree relative attention helps to find relatives in AST tree, e.g., parent and siblings, which is important in type prediction. The advantage of sequential relative attention is that it can use the multidimensional embeddings of relations: the sequential relations are shared across objects, leading to affordable 3-dimensional embedding tensors of shape (length, length, embedding dimension). In contrast, tree relative attention can only afford one-dimensional embeddings of relations, because tree-based relations are not shared between objects in a mini-batch, and extracting them for a mini-batch would already lead to a 3-dimensional tensor: (batch, length, length).

GGNN Sandwich achieves high results in the variable misuse task for which this model was developed. The reason is that in variable misuse detection, the goal is to choose two variables, and \emph{local} message passing informs each variable of its role in a program and makes variable representations more meaningful. The original work on GGNN Sandwiches also uses additional types of edges which would improve the performance of this model further. Using these types of edges is out of scope of this work, since we focus on utilizing \emph{syntactic} information, thus we only use syntax-based edges.

In Appendix~\ref{app:ano}, we visualize the progress of test metrics during training, for different structure-capturing mechanisms in the full-data setting. This Appendix also presents the comparison of structure-capturing mechanisms in the anonymized setting that is described in Section~\ref{sec:method} and implies replacing values in ASTs with unique placeholders.
The leading mechanisms are the same in all tasks as in the full data setting, considered above. An interested reader may also find examples of attention maps for different mechanisms in Appendix~\ref{app:attention_maps}.

\begin{table}[t!]
	\caption{Time- and storage-consumption of different structure-capturing mechanisms for the variable misuse task on the Python dataset.}\label{tab:performance}
	\begin{tabular}{c|ccc}
		                          & Train time    & Preprocess & Add. train  \\ 
		           Model          &  (h/epoch)  & time (ms/func.)   & data (GB)            \\ 
		 \midrule
		         Seq. pos. emb.    &  2.3    &  0  & 0                          \\ 
		      Seq. rel. att.    &  2.7    &  0  &  0                    \\
		         Tree pos. enc.    &  2.5    &  0.4  &   0.3                  \\ 
		         Tree rel. attn.   &  3.9    &  16.7  &   18   \\ 
		         GGNN Sandwich      & 7.2    &  0.3  &   0.35    \\ 
	\end{tabular}
\end{table}

\begin{table}[t!]
\caption{Comparison of combinations of sequential relative attention (SRA) with other structure-capturing approaches. All numbers in percent, standard deviations: VM: 0.5\%, FN: 0.4\%, CC: 0.1\%. Bold emphasizes combinations that significantly outperform SRA. *In the VM task, SRA+GGNN Sandwich significantly outperforms SRA during the first half of epochs, but loses superiority at the last epochs, for both datasets. On the Python dataset, SRA+GGNN Sandwich outperforms SRA by one standard deviation at the last epoch. }\label{tab:exp2b}
	\begin{tabular}{c|c|ccc}
		 & Model & VM & FN  &  CC (val.)  \\ 
		 \midrule
		PY & SRA                 & 81.42  & 35.73 & 54.53 \\ 
		\hline
		& SRA + Seq. pos. emb.      & 80.77  & 33.99 & 54.37  \\ 
		PY & SRA + Tree pos. enc.     & 81.73 &  34.71& 54.63 \\ 
		 & SRA + Tree rel. attn.    & 81.58 & 35.41  & \textbf{54.91} \\ 
		& SRA + GGNN  sand.           & 82.00*  & 33.39 & N/A  \\
		\hline
		JS & SRA         & 76.52   & 24.62  & 64.11  \\ 
		\hline
		& SRA + Seq. pos. emb.      & 73.17 & 23.09 & 63.97  \\ 
		& SRA + Tree pos. enc.     & 74.73 & 23.70 & \textbf{64.49}   \\ 
		JS & SRA + Tree rel. attn.    & 76.34 & 24.71 &  \textbf{64.79}   \\ 
		& SRA + GGNN  sand.           & 75.33* & 21.44  & N/A \\
	\end{tabular}
\end{table}

In Table~\ref{tab:performance}, we list training time and the size of auxiliary data needed for different structure-capturing mechanisms. GGNN Sandwich model requires twice the time for training (and prediction) compared to other models, because of the time-consuming message passing mechanism. Tree relative attention requires dozens of gigabytes for storing pairwise relation matrices for all training objects that could be replaced with slow on-the-fly relation matrix generation. Tree positional encodings and GGNN Sandwich models also require additional disk space for storing preprocessed graph representations, but the sizes of these files are relatively small. Sequential positional embeddings and relative attention are the most efficient models, in both time- and disk-consumption aspects. 

To sum up, \textit{we emphasise sequential relative attention as the most effective and efficient approach for capturing AST structure in Transformer}. 

\paragraph{Combining structure-capturing mechanisms.}
In Table~\ref{tab:exp2b}, we show that \textit{using sophisticated structure-capturing mechanisms
may be useful for further improving sequential relative attention if we combine two mechanisms}. We find that tree relative attention (for both datasets) and tree positional encoding (for JS) improve the score in the value prediction task, while GGNN Sandwich may improve the score in the variable misuse task, especially at earlier epochs. 

\section{Capability of Transformer to utilize syntactic information}
\label{sec:exp1}

\begin{table}[t!]
	\caption{Illustration of different kinds of models used in the experiments. The code snippet and its AST used in the illustration may be found in Figure~\ref{fig:concepts} (a, b).}\label{tab:syntaxtext}
	\begin{tabular}{l|l}
		Model & Input representation \\  \midrule
		\emph{Syntax+Text} & [(\verb|Assign|, \verb|<empty>|),
		(\verb|NameStore|, \verb|elem|), ... \\ 
		&  
	    \hspace{1cm} ...,	(\verb|Index|, \verb|<empty>|),	(\verb|NameLoad|, \verb|idx|)] \\ \hline
		\emph{Syntax} & [(\verb|Assign|, \verb|<empty>|), 
		(\verb|NameStore|, \verb|<var1>|), ...\\ 
		& \hspace{0.5cm} ...,	(\verb|Index|, \verb|<empty>|),
		(\verb|NameLoad|, \verb|<var3>|)] \\ \hline
		\emph{Text} & [\verb|elem|, \verb|lst|, \verb|idx|] \\ \hline
		\emph{Constant} & Predicts the most frequent target for any input
		\\
	\end{tabular}
	
\end{table}

\begin{figure*}[t!]
    \centering
        \begin{tabular}{ccc}
         \hspace{0.5cm}  Variable Misuse -- Python & \hspace{0.5cm} Function Naming -- Python  & \hspace{0.5cm} Code Completion (values) -- Python\\
         \includegraphics[width=0.25\linewidth]{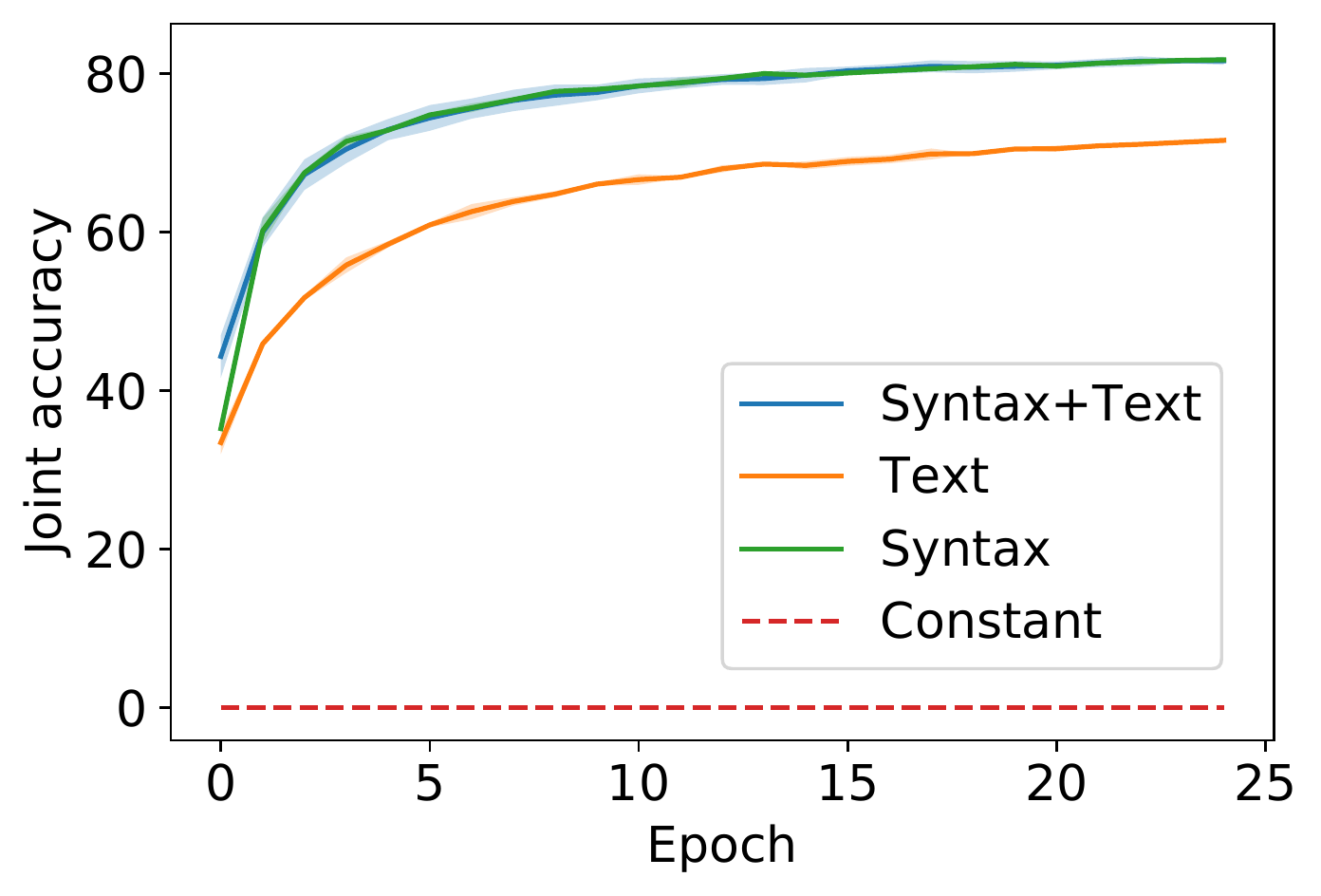} &
         \includegraphics[width=0.25\linewidth]{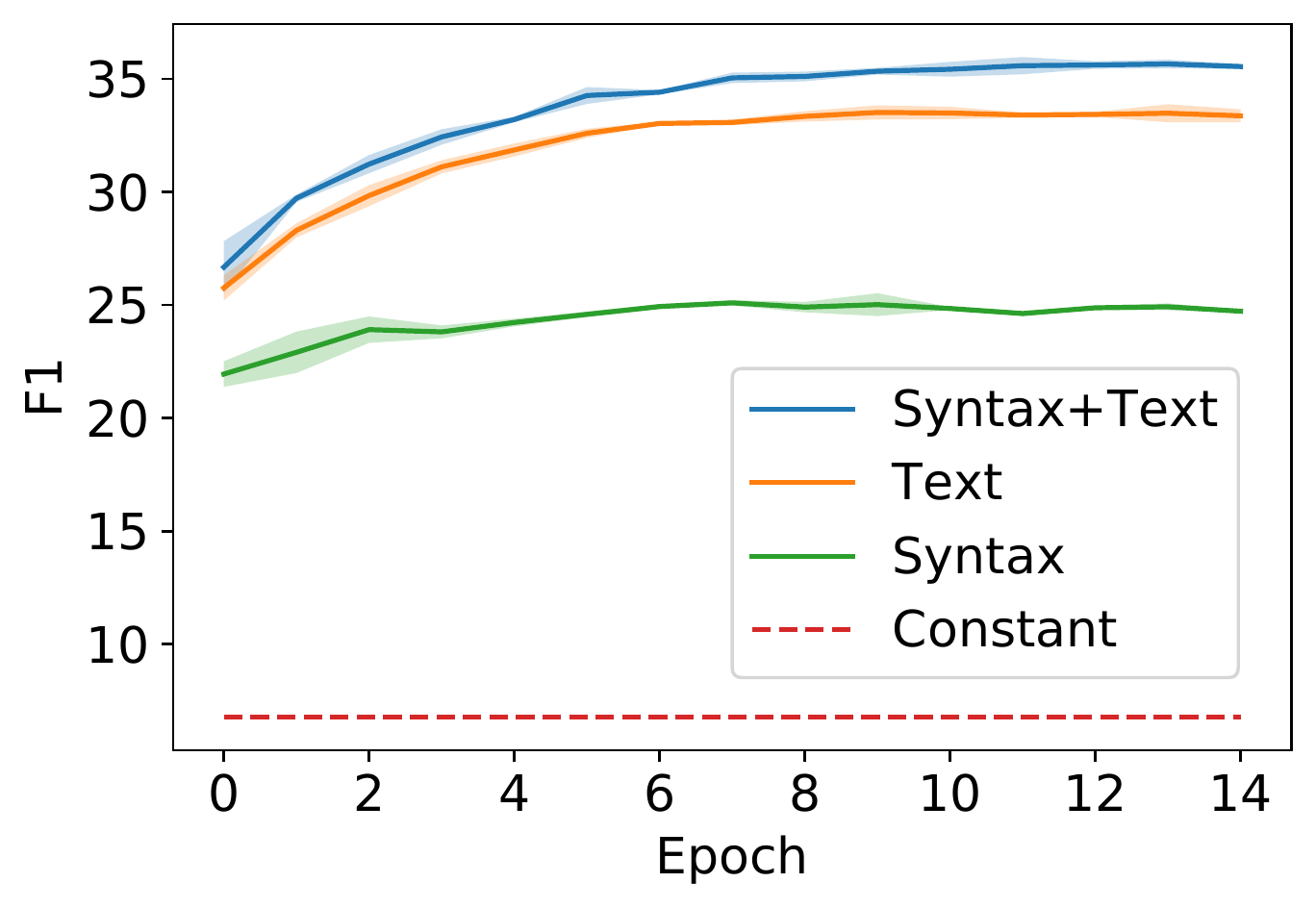} &
\includegraphics[width=0.25\linewidth]{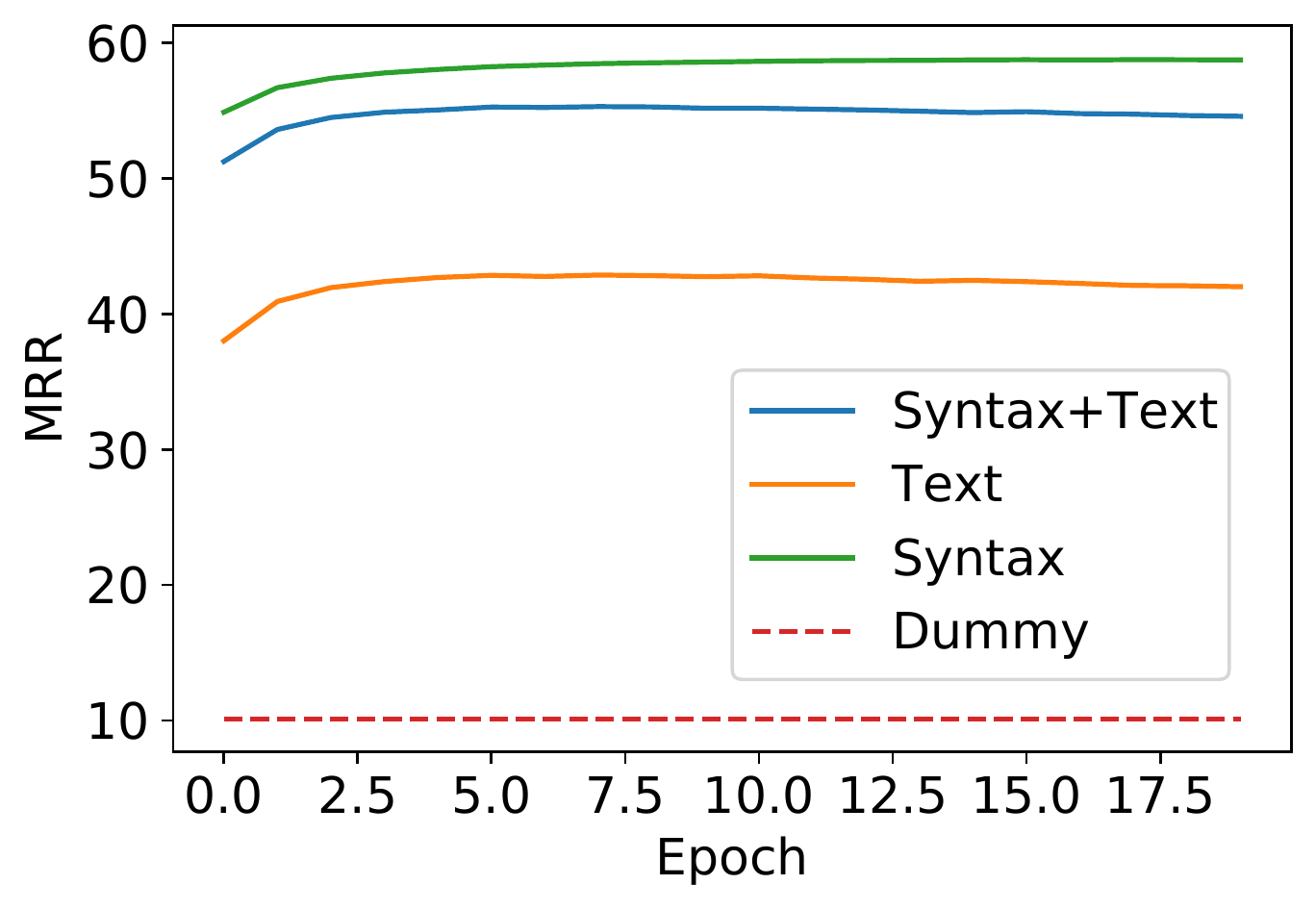} \\ 
        \hspace{0.5cm} Variable Misuse -- JavaScript & \hspace{0.5cm} Function Naming -- JavaScript & \hspace{0.5cm} Code Completion (values)  -- JavaScript \\
         \includegraphics[width=0.25\linewidth]{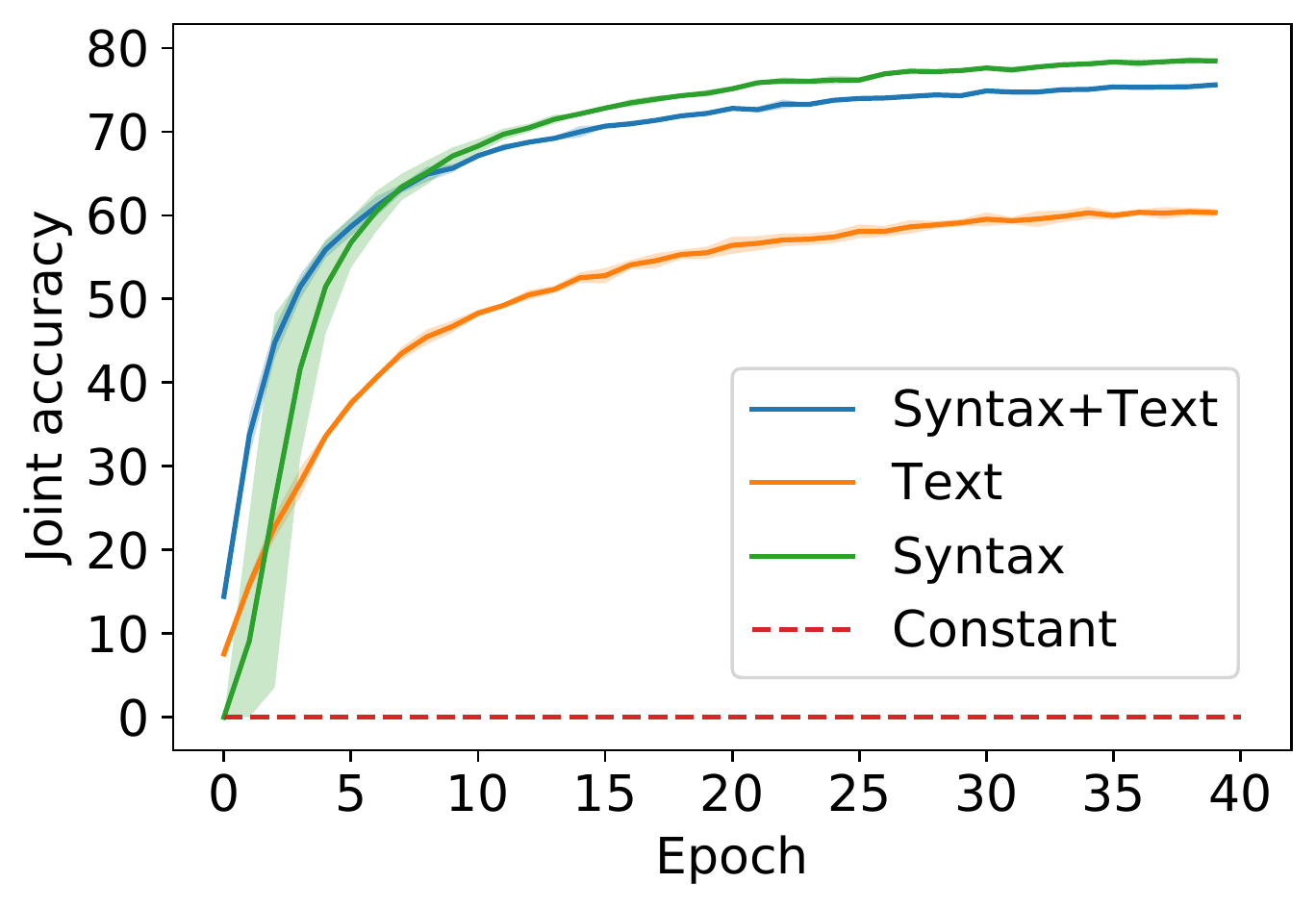} &
         \includegraphics[width=0.25\linewidth]{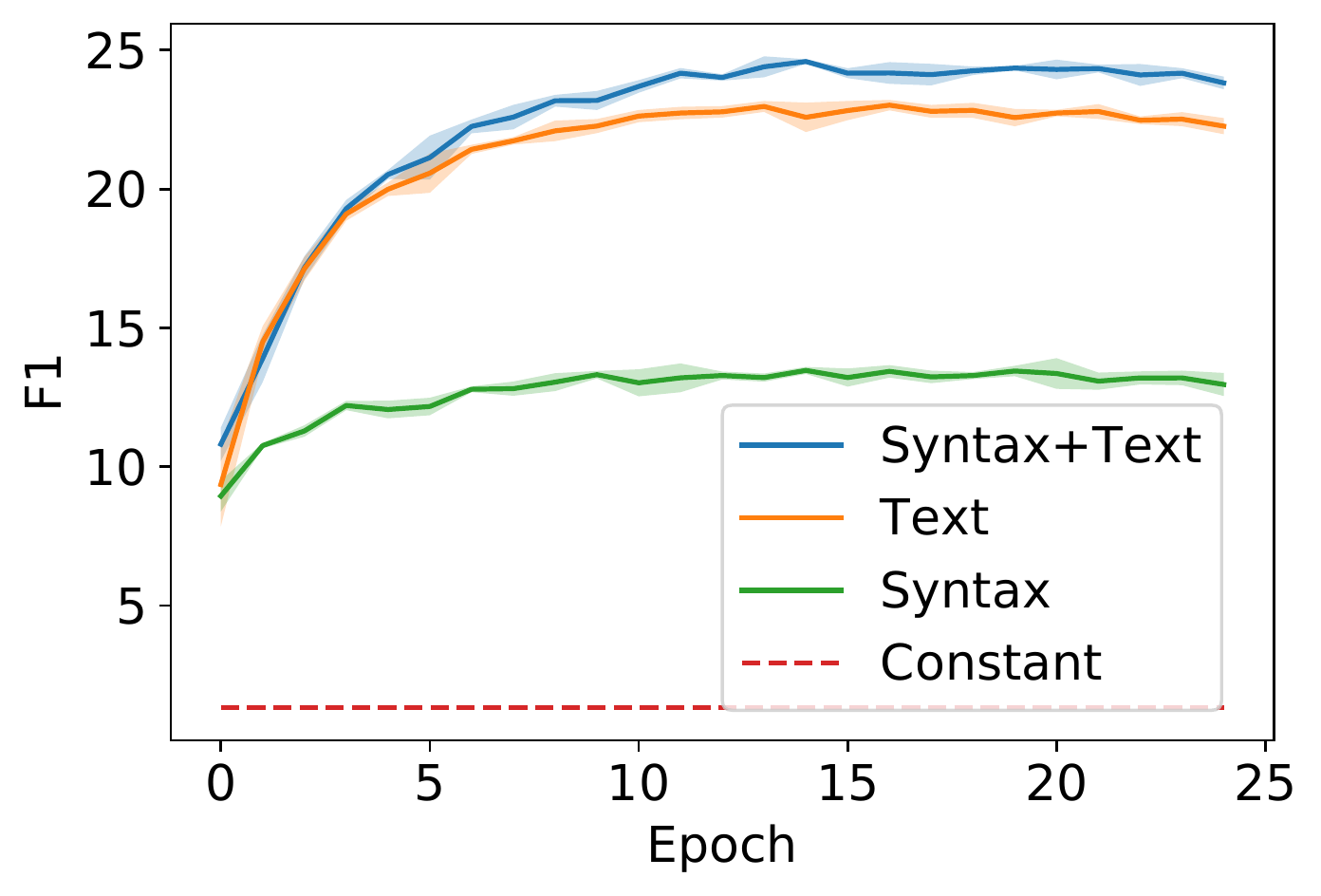} &
\includegraphics[width=0.25\linewidth]{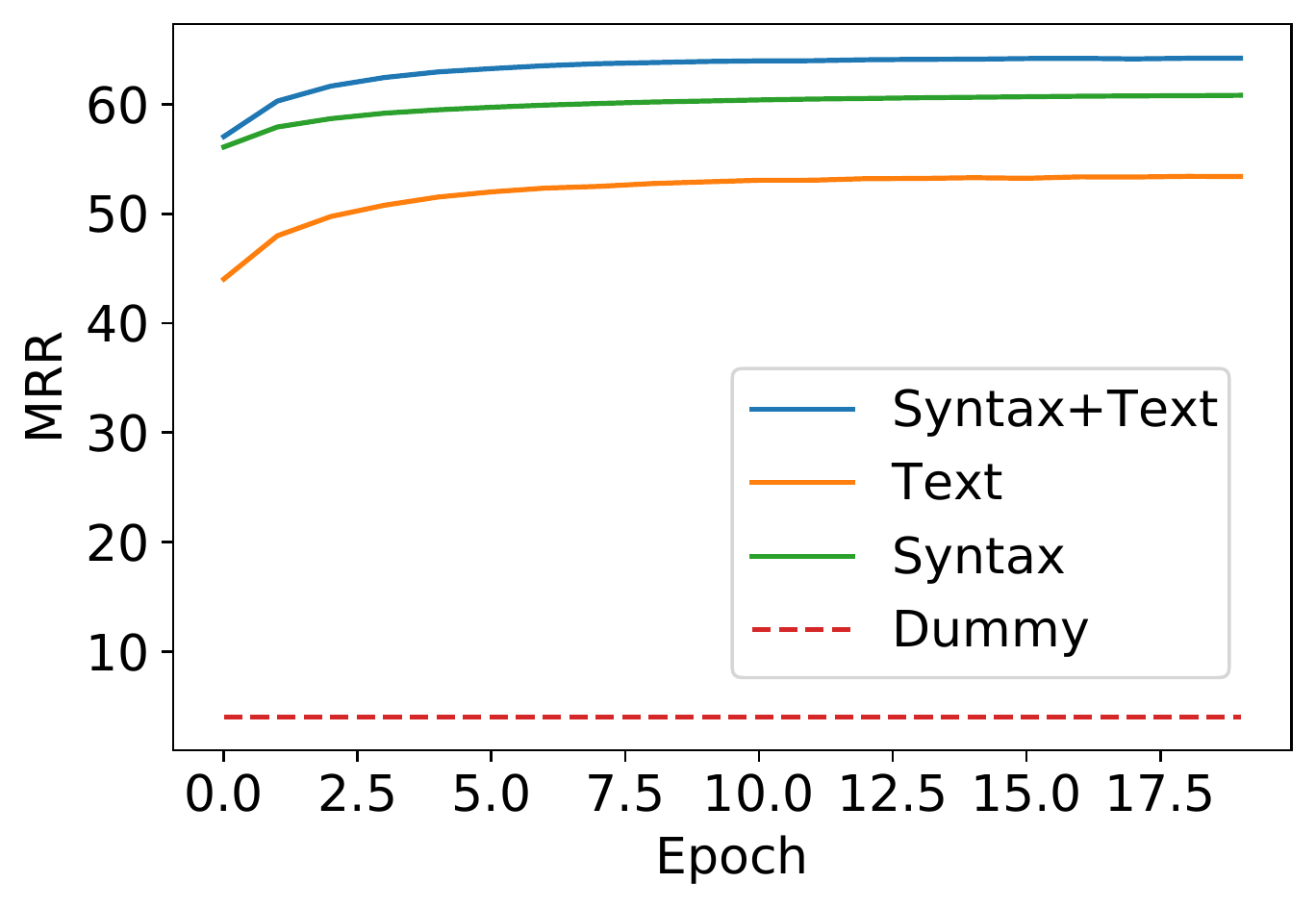}
        \end{tabular}
        \caption{
        Comparison of syntax-based Transformer models with text-only and constant baselines.
        }
        \label{fig:exp1}
\end{figure*}

\begin{figure*}[t!]
    \centering
     \includegraphics[width=0.97\linewidth]{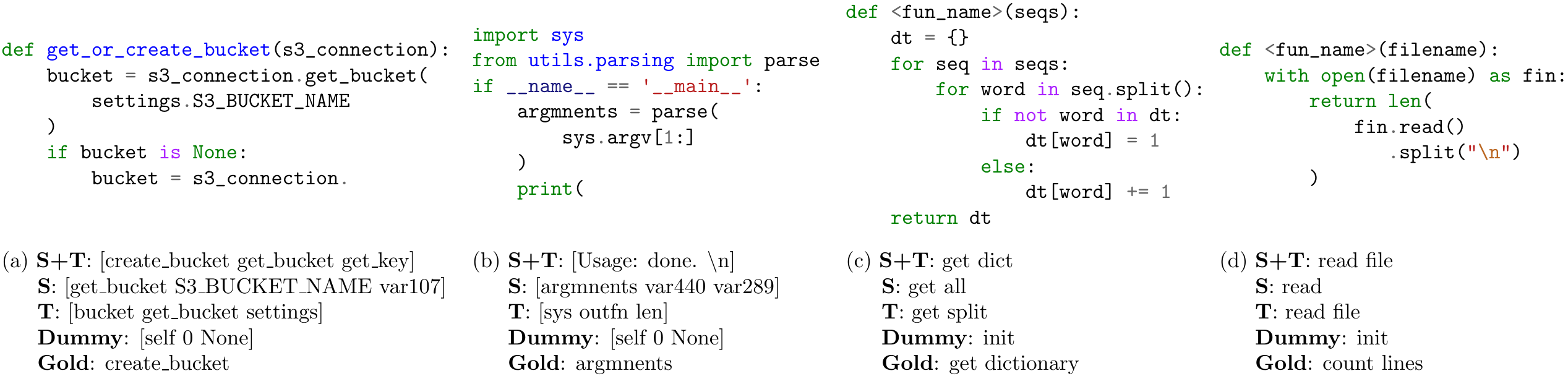}
        \caption{Example predictions in code completion (a, b; top 3 predictions) and function naming (c, d) tasks.  \textbf{S}: \textit{Syntax}, \textbf{T}: \textit{Text}.}
        \label{fig:demo}
\end{figure*}

When developing approaches for utilizing syntactic information in Transformer, the majority of works mostly focus on the tree \textit{structure}. 
We discussed the utilization of structure in the previous section, but we also would like to investigate 
the influence of other AST components, namely \textit{types} and \textit{values}. In the next section, we conduct an ablation study of the mentioned components, and in this section, we investigate
whether Transformer is generally capable of utilizing the syntactic information in source code, i.\,e.\,does processing AST components improve performance. This conceptual experiment tests the overall suitability of Transformer for source code processing. We formalize the specified question in full-data and anonymized settings as follows:
\begin{itemize}
     \item \textit{Syntax+Text} vs. \textit{Text}: First, we test whether using syntactic information in addition to the textual information is beneficial, compared to using pure textual information. To do so, we compare the quality of Transformer trained on full AST data (\emph{Syntax+Text}) with the quality of Transformer trained on a sequence of non-empty values (\emph{Text}), see Table~\ref{tab:syntaxtext} for illustrations. The \emph{Text} model relies only on textual information stored in values and does not have access to any other kind of information.
    \item \textit{Syntax} vs. \textit{Constant}: Secondly, we test whether Transformer is able to make meaningful predictions given \emph{only} syntactic information, without  textual information. To do this, we test whether the quality of Transformer trained on the \emph{anonymized} AST data (\emph{Syntax}) is better than the quality of a simple constant baseline (\emph{Constant}). Anonymization removes identifiers, i.e., textual information, but preserves information about identifiers' repetition, which may be essential for understanding the program. The \emph{Constant} model outputs the most frequent target, e.g., no-bug in VM and name \verb|init| for PY and \verb|exports| for JS in FN. Since anonymized AST data contains only syntactic information and does not contain textual information, the only way the Transformer can outperform the  \emph{Constant} baseline on this data is to capture some information from AST.
\end{itemize} 
All the described models are trained with sequential relative attention.
Deduplicating the dataset is very important in this experiment to avoid overestimating the \textit{Syntax} baseline.

The results for three tasks are presented in Figure~\ref{fig:exp1}. 
In all cases, \textit{Syntax+Text} outperforms \textit{Text}, and \textit{Syntax} outperforms \textit{Constant}. In Figure~\ref{fig:demo}, we present example predictions for code completion and function naming tasks for Python language with all four models. 
In code completion, syntax-based models capture frequent coding patterns well, for instance, in example (b), \textit{Syntax} model correctly chooses (anonymized) value \verb|argmnents| and not \verb|argv| or \verb|parse|
because \verb|argmnents| goes before assignment. 
In example (a), \textit{Syntax+Text} correctly predicts \verb|create_bucket| because it goes inside the if-statement checking whether the bucket exists, while \textit{Text} model outputs frequent tokens associated with buckets.

In the function naming task, the \textit{Text} model captures code semantics based on variable names and sometimes selects wrong ``anchor'' variables, e.g., \verb|split| in example (c), while the \textit{Syntax+Text} model utilizes AST information and outputs correct word \verb|dict|. The \textit{Syntax} model does not have access to variable names as it processes data with placeholders, and outputs overly general predictions, e.g., \verb|get all|~~in example (c). Nevertheless, the \textit{Syntax} model is capable of distinguishing general code purpose, e.g., model uses word \verb|get| in example (c) and word \verb|read| in example (d). To sum up, \textit{Transformer is indeed capable of utilizing syntactic information}. 

Interestingly, in code completion (value prediction, PY) and variable misuse detection tasks (JS), the \textit{Syntax} model, trained on the anonymized data 
outperforms the \textit{Syntax+Text} model trained on full data, though the latter uses more data than the former. The reason is that the values vocabulary on full data is limited, so approx. 25\% of values are replaced with the \verb|UNK| token and cannot be predicted correctly. On the other hand, this is not a complication for the \textit{Syntax} model, which anonymizes both frequent and rare identifiers in the same way. For example, in Figure~\ref{fig:demo}(b), the \textit{Syntax} model correctly predicts the misspelled token \verb|argmnents| while for the \textit{Syntax+Text} model, this token is out-of-vocabulary and so model outputs frequently used strings for printing. One more advantage of the \textit{Syntax} model in the value prediction task is that it is twice faster in training because of the small output softmax dimension.
In the function naming task, the \textit{Syntax} model performs substantially worse than \textit{Syntax+Text}, because variable names provide much \emph{natural language} information needed to make natural language predictions. To sum up, \textit{anonymization may lead to a higher quality than using full data}.

\section{Ablation study of mechanisms for utilizing syntactic information in Transformer}
\label{sec:exp2}

In this section, we investigate the effect of ablating different AST components on the performance in three tasks. This ablation study is important for both providing practical recommendations and understanding what mechanisms are essential for making reasonable predictions in the anonymized setting, discussed in the previous section.
We consider three AST components (comments in items regard the usual scenario without ablation): 
(\textit{types}): the types of nodes are passed as one of the Transformer inputs;
(\textit{values}): the (anonymized) values are passed as one of the Transformer inputs; (\textit{structure}): AST structure is processed using one of the mechanisms discussed in Section~\ref{sec:review}. 

As in Section~\ref{sec:exp1}, we consider both anonymized and full-data settings. We ablate AST components one by one in both models \textit{Syntax} and \textit{Syntax+Text} and check whether the quality drops.
Ablating \textit{types} for \textit{Syntax+Text} was in fact performed in Section~\ref{sec:exp1}, but we repeat the results in this experiment's table and also report this ablation for the anonymized setting. 
Ablating \textit{structure} means turning off all syntax-capturing mechanisms so that the input of the \textit{Syntax+Text}~/~\textit{Syntax} model will be viewed as an  \emph{unordered set} of (type, value)~/~(type, anonymized value) pairs.
Ablating \textit{values} means using only types as the input to the model --- the numbers are the same for both full-data and anonymized settings. We skip this ablation in the code completion task, since, in this case, (anonymized) values are the target of the model.

\begin{table*}[ht!]
		\caption{Ablation study of processing different AST components in Transformer. Bold emphasises best models and 
	ablations that do not hurt the performance. AST w/o struct.: Transformer treats input as a bag without structure; AST w/o types: only values or anonymized values are passed to Transformer; AST w/o an.val.: only types are passed to Transformer. N/A -- not applicable.}
	\label{tab:exp1b}
	
	\centering
	\begin{tabular}{c|c|ccc|ccc}
  & &	\multicolumn{3}{c|}{Full data} & \multicolumn{3}{c}{Anonymized data} \\  \midrule
	&	 & Var. misuse & Fun. naming  & Comp. (val.) &
		Var. misuse & Fun. naming  & Comp. (val.)\\ 
		\hline
	                                &	Full AST        & 
	        \textbf{81.59}$\pm$0.50\%     & \textbf{35.73}$\pm$0.19\%               &       \textbf{54.59}$\pm$0.2\%   & 
	        \textbf{81.71}$\pm$0.41\%      & \textbf{25.26}$\pm$0.15\%             &       \textbf{58.76}$\pm$0.2\% \\ 	
	Python                          &	AST w/o struct. & 
	
	        26.81$\pm$0.47\%               & 34.80$\pm$0.24\%               &       53.1$\pm$0.1\%  & 
	        12.41$\pm$0.58\%               & 23.29$\pm$0.18\%                        &       57.75$\pm$0.05\% \\ 	
	        &	AST w/o types & 
	         71.55$\pm$0.28\%               & 33.60$\pm$0.23\%               &       42.01$\pm$0.05\%  & 
	         58.55$\pm$0.51 \%               &  12.50$\pm$1.5\%                        &      41.26$\pm$0.05\% \\ 	
                            	    &	AST w/o an.val. & 
                    	
	        32.44$\pm$0.35\%                & 25.25 $\pm$0.06\%              &       N/A           &    
	       32.44$\pm$0.35\%                & \textbf{25.25}$\pm$0.06\%              &       N/A \\	\hline      

	                                &	Full AST    &
	       \textbf{75.60}$\pm$0.15\%              &    \textbf{24.62}$\pm$0.14\%           &       \textbf{64.2}$\pm$0.05
	   &    \textbf{78.47}$\pm$0.26\%                &      \textbf{13.66}$\pm$0.30\%       &       \textbf{60.82}$\pm$0.07\% \\ 	
	JavaScript  &	
	                                    AST w/o struct. & 
	                                
	      17.25$\pm$0.83\%                        & 23.40$\pm$0.12\%                       &       61.53$\pm$0.15\%   & 
	       5.37$\pm$0.97\%                        & 11.25$\pm$0.08\%                       &       58.59$\pm$0.1\% \\ 		
	                                &	AST w/o types & 
	
	         60.33$\pm$0.50\%               & 23.09$\pm$0.09\%               &     53.4$\pm$0.1 \%  & 
	         43.53$\pm$0.92 \%               &  8.10$\pm$1.4\%                        &    42.91$\pm$0.1\% \\ 	
	                               &     AST w/o an.val.     & 
	                                
	       42.56$\pm$0.24\%                       & 13.64$\pm$0.07\%              &       N/A                       &  
	       42.56$\pm$0.24\%                       & \textbf{13.64}$\pm$0.07\%              &       N/A \\                      
	\end{tabular}

\end{table*}


The results are presented in Table~\ref{tab:exp1b}. \textit{In variable misuse and code completion, all AST components are essential for achieving high quality results}. Particularly, in variable misuse, all ablations result in a large quality drop, in both settings, and in code completion, ablating types results in a large quality drop and ablating structure --- in substantial drop. Interestingly, anonymization plays an important role in achieving high quality in the variable misuse task, with absent values from the data.

However, the observations differ for function naming. In this task, (1) ablating \textit{types} results in a substantial quality drop in both settings, (2) ablating \textit{structure} results in a small (but significant) quality drop in both settings, (3) ablating \textit{values} in the full-data setting results in the large quality drop, and (4) ablating \textit{anonymized values} does not affect the performance in the anonymized setting. The first and the third observations underline the importance of using both types and values in practice. The second and the fourth observations show that \textit{Transformer is now far from utilizing all the information stored in AST when predicting function names}. Particularly, in the anonymized setting, Transformer predicts function names mostly based on \textit{types}. It hardly uses syntactic \textit{structure}, and does not use information about value repetition which is stored in \textit{anonymized values} and is essential for understanding the algorithm that the code implements. Overcoming this issue is an interesting direction for future research.

\section{Ensembling of syntax-based models}
\label{sec:exp3}

\begin{table}[t!]
\caption{Comparison of ensembles. Notation: ST -- \emph{Syntax+Text}, S -- \emph{Syntax}, \& denotes ensembling. All models are trained with sequential relative attention. All numbers in percent, standard deviations: VM: 0.5\%, FN: 0.4\%, CC: 0.1\%.}\label{tab:ensembling}
	\begin{tabular}{c|c|cccc}
		& Models & VM & FN  & CC (types) & CC (values) \\ 
		 \midrule
		& ST                & 81.42    & \textbf{35.73} & 89.22             & 54.53  \\
		& ST \& ST          & 82.80    & \textbf{35.61}          & \textbf{89.39}             & 56.35  \\
		PY & S              & 81.83    & 25.26          & 88.42             & 58.6  \\
		 & S \& S           & 82.57    & 25.46          & 88.65             & 59.29 \\
		& ST \& S    & \textbf{86.72}   & 32.15         & \textbf{89.49}             & \textbf{61.84}  \\
		\hline
		& ST              & 76.52 &  \textbf{24.62}         &  90.14    &   64.11  \\
		& ST \& ST            &  77.25   & \textbf{24.53}            &  \textbf{90.56}     &  65.68  \\
		JS & S              &   78.53  &  13.66             &  88.22    & 60.71   \\
	    & S \& S            &  79.65   & 13.19              &  88.52    &  61.42  \\
		& ST \& S    & \textbf{82.29}  &  19.33             &  90.32    &  \textbf{68.33}  \\
	\end{tabular}
\end{table}

As was shown in Figure~\ref{fig:demo}, \textit{Syntax+Text} and \textit{Syntax} models capture dependencies of different nature and are orthogonal in a sense of handling missing values and first time seen tokens. This allows hypothesizing that \emph{ensembling} two mentioned models can boost the performance of the Transformer.  
We use the standard ensembling approach that implies training  networks from different random initializations and averaging their predictions after softmax~\cite{pitfalls}. We use sequential relative attention in this experiment.

In Table~\ref{tab:ensembling}, we compare an ensemble of \textit{Syntax+Text} and \textit{Syntax} models with ensembles of two \textit{Syntax+Text} and of two \textit{Syntax} models. We observe that in variable misuse and value prediction tasks, ensembling models that view input data in two completely different formats is much more effective than ensembling two similar models. This is the way how using anonymized data may boost the Transformer's performance.

\section{Validating our implementations and comparing to other works}
\label{sec:valid}
We ensure the validity of our results in two ways: by relying on the code of already published works, and by comparing our numbers achieved for the commonly used data split to the numbers in the corresponding papers. Particularly, we use the model / loss / metrics / overlapping chunks code of \citet{code-prediction-transformer} as the baseline for the CC task, we rewrite (line by line) the main parts of the model / loss / metrics code of \citet{Hellendoorn} in PyTorch, as the baseline for the VM task, and we use the model / loss / metrics code of \citet{summarisation} as the baseline for the FN task. 

For VM, the vanilla Transformer of \citet{Hellendoorn} achieve 67.7\% joint accuracy and we achieve 64.4\%: the results are close to each other. Here the performance is given for our model, closest to the model of~\citet{Hellendoorn}: the \textit{Text} model of similar size and with similar number of training updates; using our \textit{Syntax+Text} model achieves higher quality. The performance of GGNN Sandwich is high on VM, as in~\cite{Hellendoorn}.
For CC, with tree relative attention, we achieve 59.79 / 91.65 MRR (values / types) while \citet{code-prediction-transformer} achieved 58.8 / 91.9 (their “TravTrans variant”); and for standard Transformer (no structure information), we achieve 59.66 / 89.16 MRR while \citet{code-prediction-transformer}  achieved 58.0 / 87.3 (their “TravTrans”) respectively, again the results are close. 
For FN, we used the code of \citet{summarisation} with our custom data and targets, so the results are not comparable, but we checked that their code produces the same numbers on their data as in the paper. 


The results given in our paper are for our custom data split and thus are not directly comparable to the numbers in other works. We argue that data resplitting is crucial for achieving correct results, see details in Section~\ref{sec:method}. At the same time, the remaining experimental setup (e. g. architecture, metrics) is the same as in recent works.

\section{Related Work}
\label{sec:litreview}
\paragraph{Variable misuse.} The field of automated program repair includes a lot of different tasks, see~\cite{repair_review} for a review, we focus on a particular variable misuse detection task. This task was introduced by \citet{allamanis_iclr2018} who proposed using GGNN with different types of edges to predict the true variable name for each name placeholder. \citet{varmisuse} enhances the VM task by learning to jointly classify, localize bug and repair the code snippet. They use an RNN equipped with two pointers that locate and fix the bug. \citet{Hellendoorn} improved the performance on VM task, using Transformers, GREAT model, and GGNN Sandwich model. 

\paragraph{Code summarization.} The task of code summarization is formalized in literature in different ways: given a code snippet, predict the docstring~\cite{deepcom}, the function name~\cite{allamanis2016convolutional}, or the accompanying comment~\cite{CodeNN}. 
\citet{allamanis2016convolutional} propose using convolutional neural networks for generating human readable function names, while
\citet{CodeNN} 
proposed using LSTM~\cite{lstm} with attention to generate natural language summaries. \citet{alon2018codeseq} proposed sampling random AST paths and encoding them with bidirectional LSTM to produce natural method names and summaries of the code. 
\citet{fernandes2018structured} proposed combining RNNs/Transformers with GGNN. 
\citet{summarisation} empirically investigate Transformers for code summarization, showing that Transformer with sequential relative attention outperforms Transformer equipped with positional encodings as well as a wide range of other models, e,\,g.RNNs. 
 
\paragraph{Code completion.}
Early works on code generation built probabilistic models over the grammar rules. \citet{pmlr-v32-maddison14} learned Markov Decision Process over free context grammars, utilizing AST. \citet{deep3} learned decision trees predicting AST nodes. \citet{treegen} generated code by expanding AST nodes, using 
natural language comments as additional source of information. \citet{completion_pointer} used LSTM with a pointer, this model  either generates the next token from vocabulary or copies the token from a previously seen position. \citet{code-prediction-transformer} proposed using Transformers for code generation, enhanced them with tree relative attention and showed that the resulting model significantly outperforms RNN-based models as well as other models~\cite{deep3}.

\paragraph{Recent advances in neural source code processing.} The recent line of work is dedicated to learning contextual embeddings for code on the basis of  Bidirectional Encoder Representations from Transformers (BERT)~\cite{bert}. Such models are firstly pretrained on large datasets providing high-quality embeddings of code, and then fine-tuned on small datasets for downstream tasks~\cite{cubert, codebert, guo2021graphcodebert}. 
All these Transformer-based models treat code as text and can potentially benefit from the further utilization of the syntactic information. Another line of research regards making Transformers more time- and memory-efficient~\cite{effTr}. Investigating the applicability of such methods to syntax-based Transformers is an interesting direction for future research.

\paragraph{Investigating neural networks for code with omitted variable names.} 
A few of previous works considered training neural networks with omitted variable names: \citet{deepfix, commitgen} trained RNNs on the data with anonymized variables, \citet{student_code} replaced variables with their types. \citet{subroutines} investigated the effect of replacing all values in the AST traversal with \verb|<unk>| value in the code summarization task, and concluded that the quality of an RNN trained on such data is extremely low. Their result aligns with ours, while we consider a more general procedure of value anonymization (that saves information about value repetition) and investigate the effect of using anonymization in a wider set of tasks for a Transformer architecture.

\section{Threats to validity}
\label{sec:threats}
We did our best to make out comparison of different AST processing mechanisms as fair as possible. However, the following factors could potentially affect the validity of our results: using the same training hyperparameters for all models, not using subtokenization, and not using data- and control-flow edges in GGNN Sandwich.
The decision not to use subtokenization was explained in Section~\ref{sec:method}. Moreover, we underline that sequential relative attention, our best performing mechanism, allows for easy combination with any subtokenization technique which will result in further quality improvement. On the contrary, this is not the case for more complex considered AST-processing mechanisms. The decision not use control- and data-flow edges was explained in Section~\ref{sec:review}. We note that adding data- and control-flow edges to the GGNN Sandwich equipped with sequential relative attention would increase the quality of this combined model even further.
As for the training hyperparameters, tuning them for each model individually would be very expensive given our limited computational resources. However, we note that our models differ only in the AST-processing mechanism that is a relatively small change to the architecture. Thus we assume that using the same training hyperparameters for different models is permissible in our work.

\section{Conclusion}
\label{sec:concl}
In this work, we investigated the capabilities of Transformer to utilize syntactic information in source code processing. Our study underlined the following practical conclusions:
\begin{itemize}
    \item sequential relative attention is a simple, fast and not considered as the baseline in previous works mechanism that performs best in 3 out of 4 tasks (in some cases, similarly to other slower mechanisms);
    \item combining sequential relative attention with GGNN Sandwich in the variable misuse task and with tree relative attention or tree positional encoding in the code completion task may further improve quality;
    \item omitting types, values or edges in ASTs hurts performance;
    \item ensembling Transformer trained on the full-data with Transformer trained on the anonymized data outperforms the ensemble of Transformers trained on the same kind of data.
\end{itemize}
Further, our study highlighted two conceptual insights. On the one hand, Transformers are generally capable of utilizing syntactic information in source code, despite they were initially developed for NLP, i. e. processing sequences. On the other hand, Transformers utilize syntactic information fully not in all tasks: in variable misuse and code completion, Transformer uses all AST components, while in function naming, Transformer mostly relies on a set of types and values used in the program, hardly utilizing syntactic structure.

\begin{acks}
We would like to thank Ildus Sadrtdinov, Ivan Rubachev and the anonymous reviewers for the valuable feedback. The results presented in Sections~\ref{sec:exp2a} and \ref{sec:exp3}
were supported by the Russian Science Foundation grant \textnumero 19-71-30020.
The results presented in Sections~\ref{sec:exp1} and~\ref{sec:exp2}
were supported by Samsung Research, Samsung Electronics. The research was supported in part through the computational resources of HPC facilities at NRU HSE.
\end{acks}

\newpage
\bibliographystyle{ACM-Reference-Format}
\balance
\bibliography{sample-base}


\clearpage
\newpage
\appendix
\section{Additional experimental details}
\label{app:hypers}
\subsection{Hyperparameter tuning}
To ensure the fairness of comparing different mechanisms for processing AST in Transformer, we tune their hyperparameters individually for each task--dataset combination.

Sequential positional embeddings do not have hyperparameters. For sequential relative attention, we tune the maximum relative distance between the elements of the sequence. We consider options $[8, 32, 128, 250]$: from small distance to the maximum input length.

For tree positional encoding, we tune the maximum path width $n_w$ and maximum path depth $n_d$. Since the embedding size is fixed and equal to $d_{\mathrm{model}}=512$, and the number of learnable parameters in the tree positional encoding equals $d_{\mathrm{model}}/(n_w \cdot n_d)$ (see details in~\cite{tree_encoding}), the greater values of $n_w$ and $n_d$ we use, the less number of the parameters in the encoding layer we have. We selected options to cover both cases when we have wide and deep paths but only one parameter as well as narrow and shallow paths but more parameters.

For tree relative attention, we tune the size of the relations vocabulary. All relations follow the pattern ``$k_u$ nodes up and $k_d$ nodes down'', $k_u, k_d = 0, 1, 2, \dots$. We select the top of the most frequent relations in the dataset. We consider options from relatively small vocabulary to the full vocabulary of relations.

For GGNN Sandwich model, we consider 6-layer and 12-layer configurations of alternating Transformer (T) and GGNN (G) layers, we also consider placing both types of layers first i.\,e.\,[T, G, T, G, T, G] or [G, T, G, T, G, T] (and similarly for 12 layers). GGNN layers include 4 message passes. We also consider omitting edges of types \verb|Left| and \verb|Right| (see Figure~\ref{fig:concepts}(f)).

The results for hyperparameter tuning are given in Table~\ref{tab:hypers}. Each model was trained once, but since we have several choices for hyperparameters, we could analyse whether there are stable dependencies or the difference in quality is a result of noise. The choice of the maximum relative distance is stable across datasets in all tasks: in the VM task, distance equal to 8 is chosen, which is interpretable since in this task, local dependencies matter much; in contrast, in the FN task, small values of maximum distance perform poorly, and in this task, global dependencies are important. The similar reasoning applies to tree positional encoding: in the ``local'' VM task, small values of ($n_w$, $n_d$) are chosen, while in the ``global'' FN task, large values outperform (4, 8) combination. For tree relative attention, changing the relation vocabulary size does not affect quality much: rare relations' embeddings are updated only several times during training, so including them does not improve the performance. For the GGNN Sandwich, the choice of the optimal hyperparameters is again stable across datasets. In the VM task, the bigger the model, the higher the quality: deep 12-layer models substantially outperform 6-layers ones, and 3-edge-type models outperform 3-edge-type models. In contrast, in the FN task, the 12-layer model fails to train properly, while the 6-layer model which achieves good quality.

\subsection{Anonymization procedure.}
We anonymize values so that inside one code snippet, different values are mapped to different placeholders \verb|var1|, \verb|var2|, \verb|var3| etc. We use random strategy for assigning placeholders: we fix the vocabulary of 1000 placeholders and for unique each value in each code snippet, choose a placeholder randomly. For example, the code snippet \verb|total_cnts[i][w] = title_cnt[i][w] + text_cnt[i][w]|  may be anonymized as follows:
\verb|var341[var30][var89] = | \newline \verb| var785[var30][var89] + var453[var30][var89]|.
Anonymization of different code snippets is independent:  one value may be replaced with different placeholders in different code snippets. For example, value \verb|total_cnt| may be replaced with \verb|var341| in one code snippet and with \verb|var110| --- in another one.

\begin{table*}[h]
	\centering
	\caption{The results of hyperparameter tuning for different structure-capturing mechanisms, tasks and datasets. The quality is measured over the validation dataset (~10\% of the training data). Variable Misuse: joint localization and repair accuracy set, Function Naming: F1-measure; Code completion (predicting values): MRR, all numbers in percent. Datasets: PY --- Python150k, JS -- JavaScript150k. Notation: Y -- Yes, N -- No, all -- the full vocabulary of relations is used. Bold font emphasizes the maximum among values in a column.}
	\begin{tabular}{|c|ccc|ccc|ccc|}
	\hline
	& & \multicolumn{2}{c|}{\textbf{Variable misuse}} & & \multicolumn{2}{c|}{\textbf{Function naming}} & & \multicolumn{2}{c|}{\textbf{Code completion}} \\  \midrule
	\textbf{Model (hyperparams.)} & & \textbf{PY} & \textbf{JS} & & \textbf{PY} & \textbf{JS} & & \textbf{PY} & \textbf{JS} \\ \hline
	Sequential                  & 8 & \textbf{81.69} & \textbf{76.98} & 8 & 32.63 & 22.36 & 8 & 52.68 & 61.36 \\
	relative attention          & 32 &  81.56 & 75.83 &  32   & 33.05 & 23.43 & 32 & \textbf{53.04} &  \textbf{61.73} \\
    (max. relative distance)    & 128 & 80.94 & 75.52 &   128  & 33.43 & 23.43 & 128  & 53.04 & 61.66 \\
                                & 250 & 81.05 & 74.75 &   250  & \textbf{33.54} & \textbf{23.76} & 250 & 52.95 & 61.64 \\ \hline
                                
    Tree positional     & 2, 64 &  64.74  & 55.06  & 2, 64 & 32.96 & \textbf{23.72} & 2, 64  & 50.89 & 54.49 \\
	encoding            & 4, 8 &  73.77 & \textbf{62.14} & 4, 8 & 30.75 & 21.00 & 4, 8 & 52.22 & 59.3 \\
    (max. path width,    &  8, 16  & \textbf{74.53} &  60.19 & 8, 16  &  33.06  & 22.29 & 8, 16 & 52.16 & 60.23 \\
     max. path depth)  &  16, 8  & 74.29 &  59.21 &  16, 8 & \textbf{33.48}  & 22.41 & 16, 8 & \textbf{52.38} & 60.32 \\ 
      &  16, 32  & 66.78 &  45.67 & 16, 32 &  32.85 & 23.02 & 16, 32 & 52.26 & \textbf{60.42} \\ \hline
      
    Tree relative & 100  &  \textbf{71.69}  & 65.55  & 100 & 33.38 & \textbf{24.25} & 10 & 51.90 &  60.32 \\
    attention &  600   & 71.31  & \textbf{66.61}  & 600  & \textbf{33.71} & 23.78 & 100 & 52.54 & 60.94 \\
    (relations vocabulary size) &   1500  &  71.52  &  64.38 &  1500 & 33.45 & 23.67 & 1000 & \textbf{52.66} & \textbf{61.19} \\
       &  all   & 71.36   & 66.20 & all & 32.18 & 23.95 & all & 52.46 & 61.05 \\ \hline
    GGNN Sandwich & 6, 2, N  &  70.39  &  65.25 & 6, 2, N & 33.04 & 23.36 & & & \\
    (the number of layers, &  6, 2, Y   & 70.25  &  65.87  & 6, 2, Y & \textbf{33.61} & \textbf{24.05} & & & \\
      the number of edge types, &  6, 3, N  &  78.16  &  68.74   & 6, 3, N & 32.72 & 21.40 & & & \\ 
      is GGNN the first layer?) &  6, 3, Y  & 78.33  &  71.06  & 6, 3, Y & 31.79 & 21.49 &  & N/A & \\
      &  12, 2, N   &  71.96  & 68.99 & 12, 2, N & 5.37  & 5.11 & &  & \\
        &  12, 2, Y   &  72.60  & 68.30 & 12, 2, Y & 9.85 & 5.11  & & & \\
       &   12, 3, N  &  \textbf{80.71}  & \textbf{73.88} & 12, 3, N & 21.26 & 3.40  & & & \\
       & 12, 3, Y   &  80.61   & 73.09  & 12, 3, Y & 20.36 & 6.27  & & & \\ \hline
	\end{tabular}
	\label{tab:hypers}
\end{table*}

\section{Comparison of approaches for utilizing syntactic structure in Transformer in the anonymized setting}
\label{app:ano}
In this section, we compare mechanisms for utilizing syntactic structure in Transformer in the anonymized setting, i.\,e.\,when mechanisms are incorporated in the \emph{Syntax} model described in~Section~\ref{sec:exp1}. Due to the high computational cost of hyperparameter tuning, in this experiment, we use the same hyperparameters as in the full-data setting. In Section~\ref{app:hypers} we discussed that the selected hyperparameters are quite interpretable in different tasks, which allows to hypothesise that in the anonymized setting, the optimal hyperparameters would be the same.

\begin{figure*}[ht!]
\begin{tabular}{cccc}
         \hspace{0.2cm} Python: \hspace{0.55cm} Variable Misuse &  Function Naming &  Code Completion (values) & Code Completion (types)\\
          \includegraphics[height=0.2\linewidth]{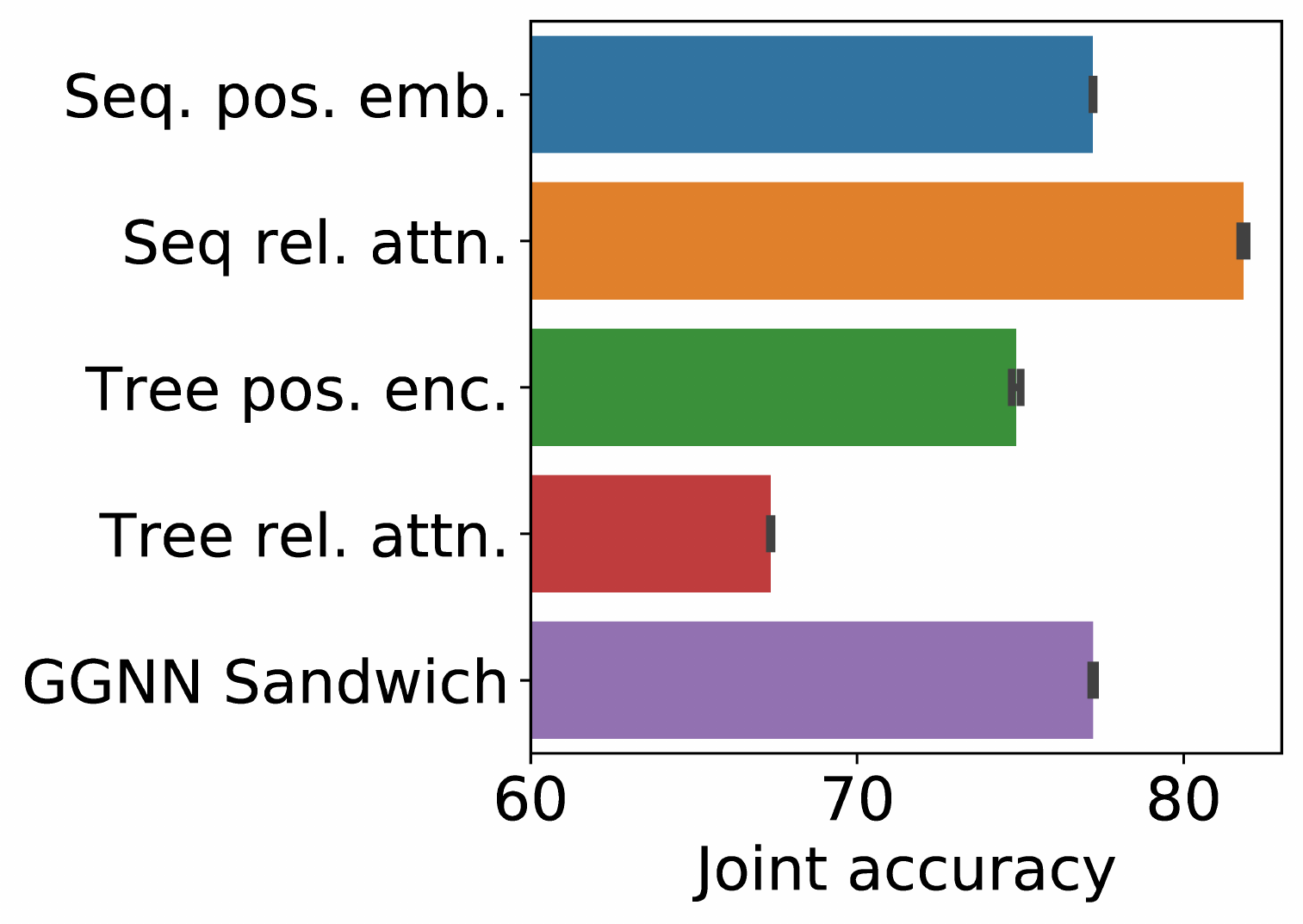} &
         \includegraphics[height=0.2\linewidth]{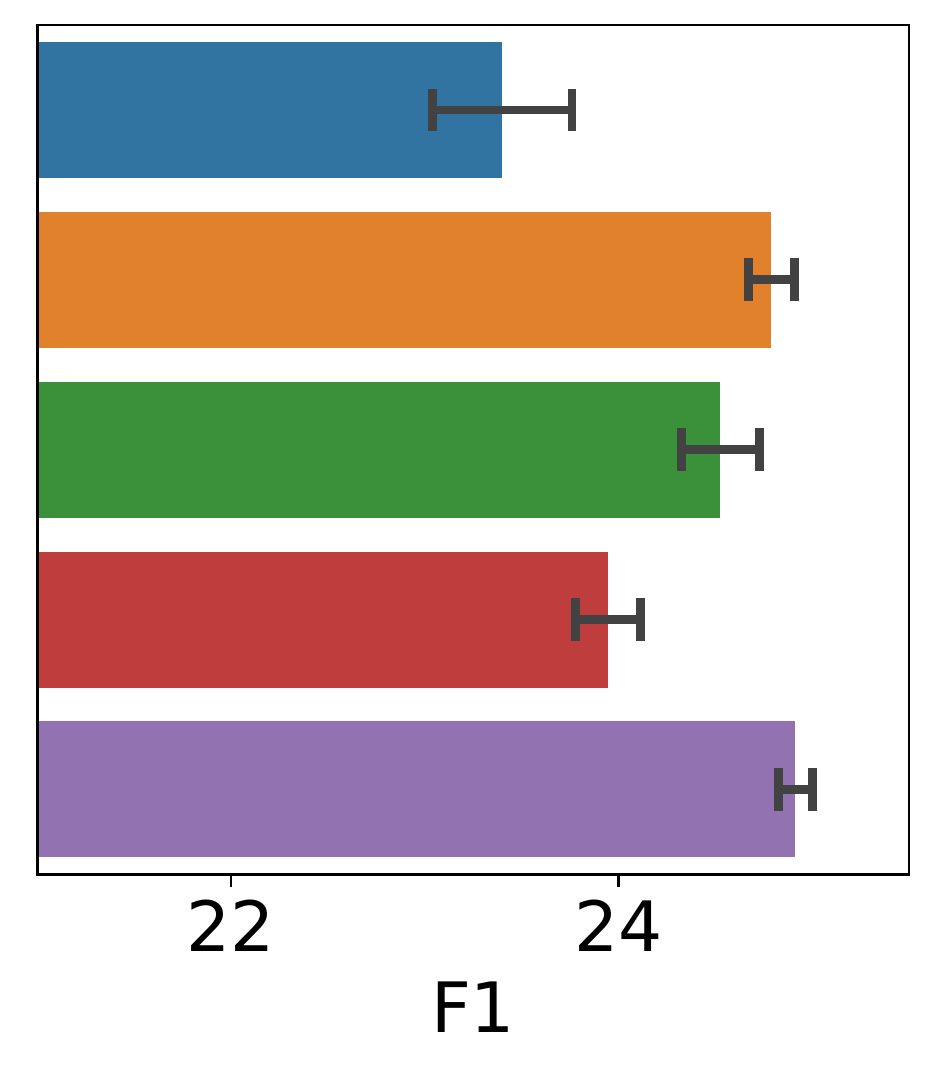} &
\includegraphics[height=0.2\linewidth]{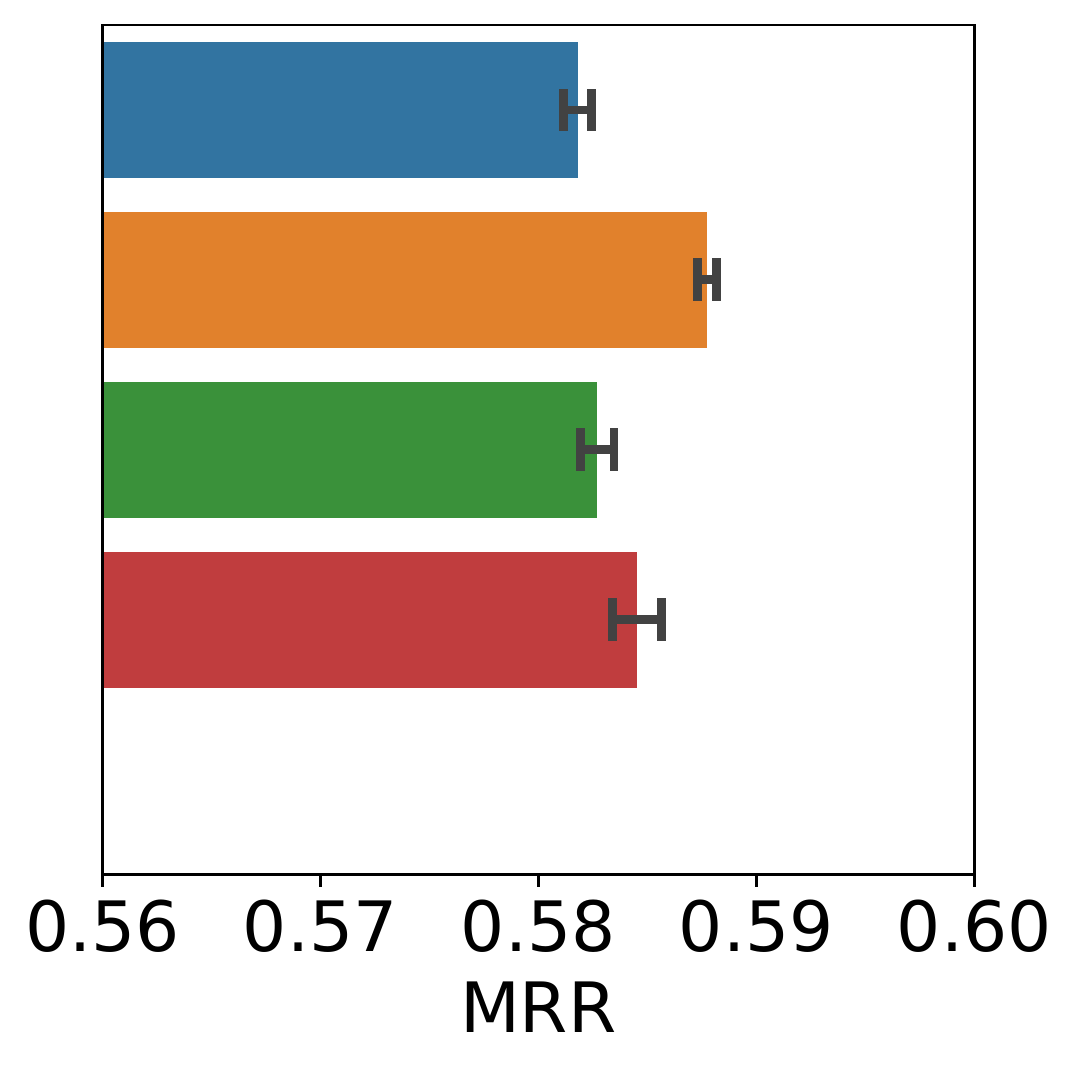}& 
\includegraphics[height=0.2\linewidth]{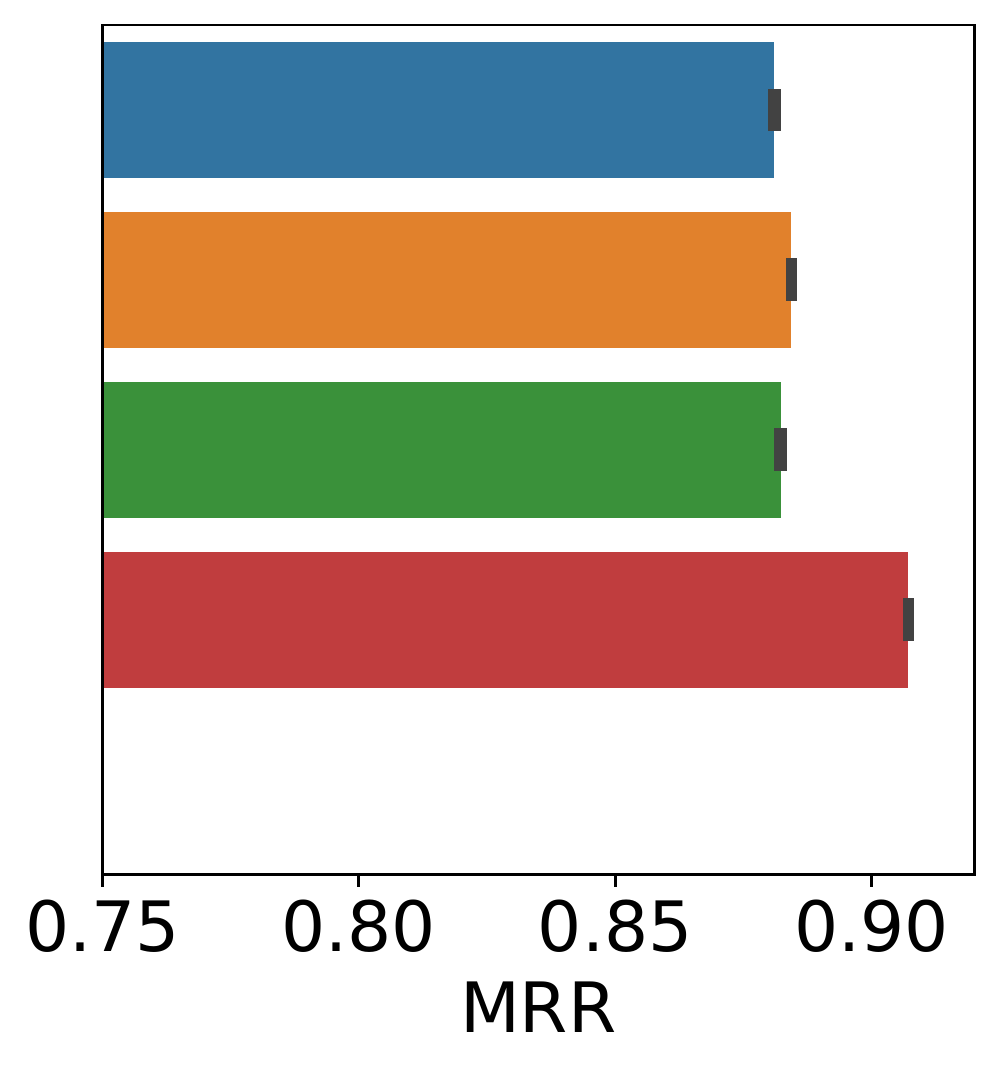}
\\
         JavaScript: \hspace{0.4cm} Variable Misuse \hspace{0.1cm} &  Function Naming  &  Code Completion (values)   &  Code Completion (types)  \\
         \includegraphics[height=0.2\linewidth]{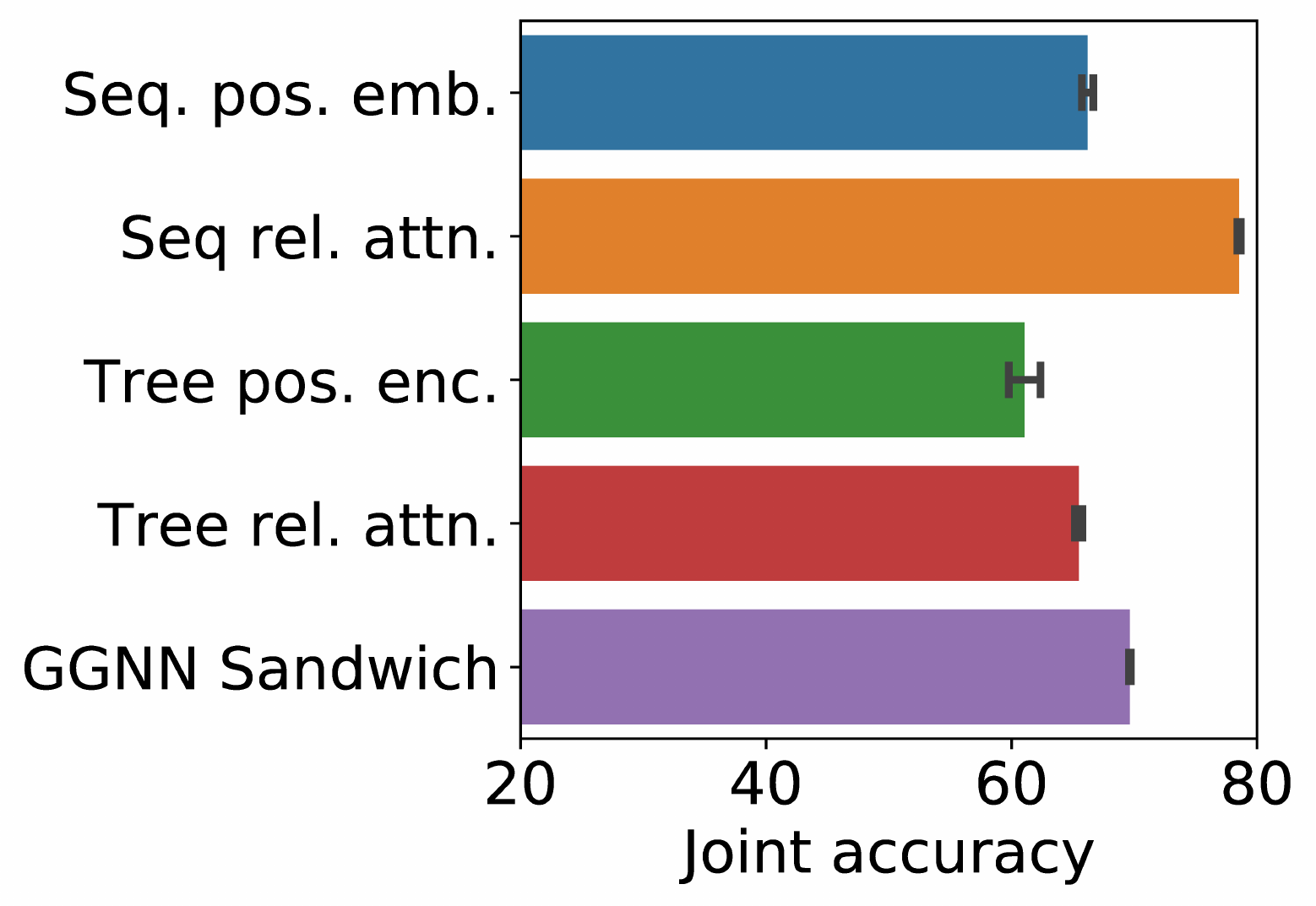} &
         \includegraphics[height=0.2\linewidth]{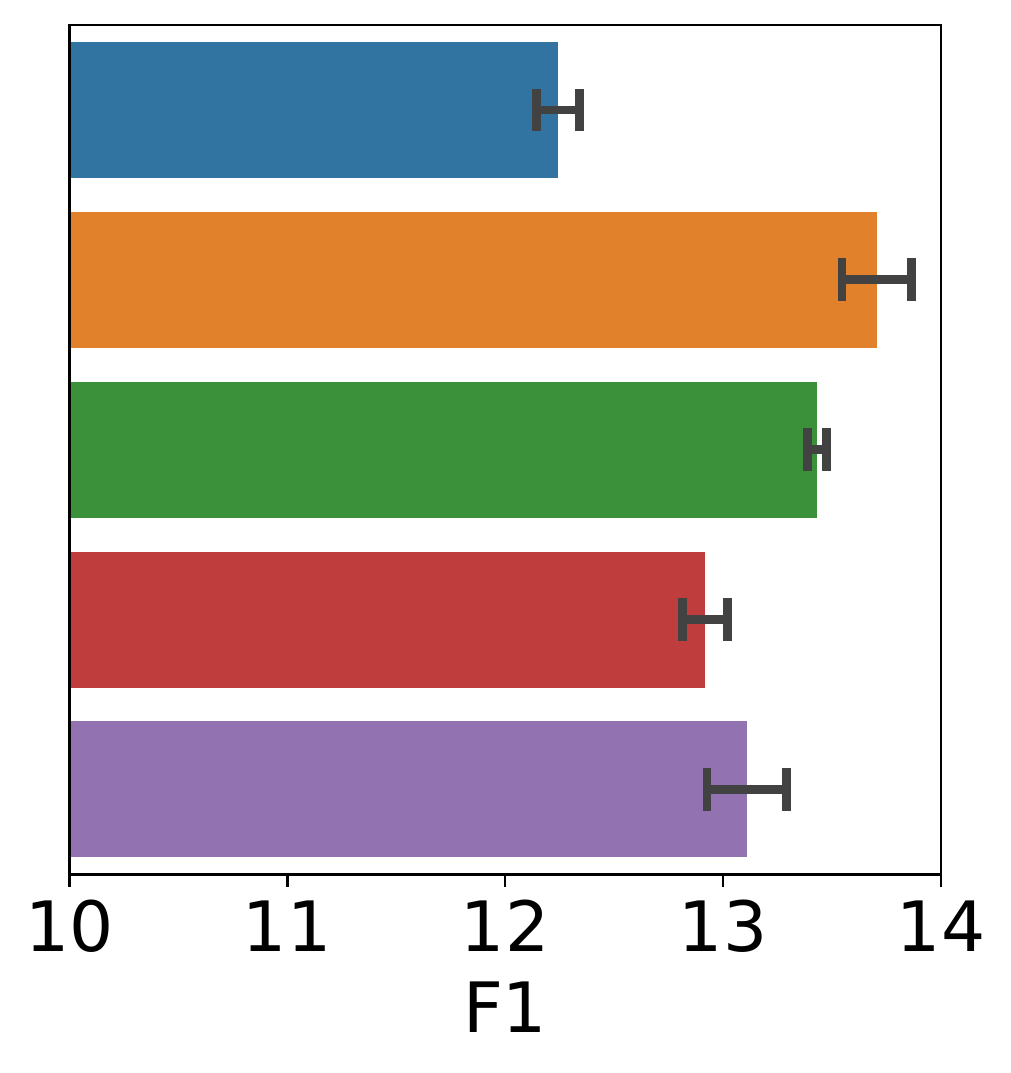} &
         \includegraphics[height=0.2\linewidth]{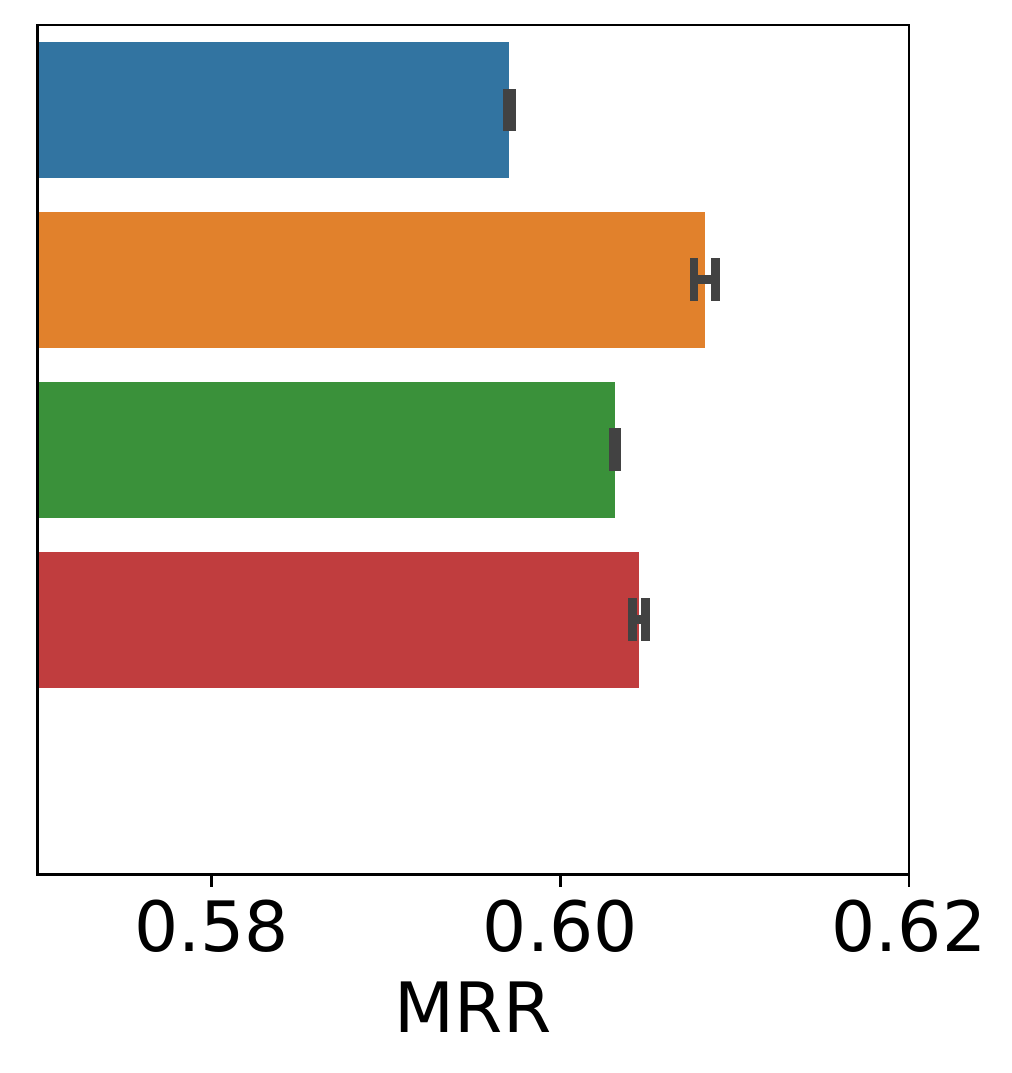} &
         \includegraphics[height=0.2\linewidth]{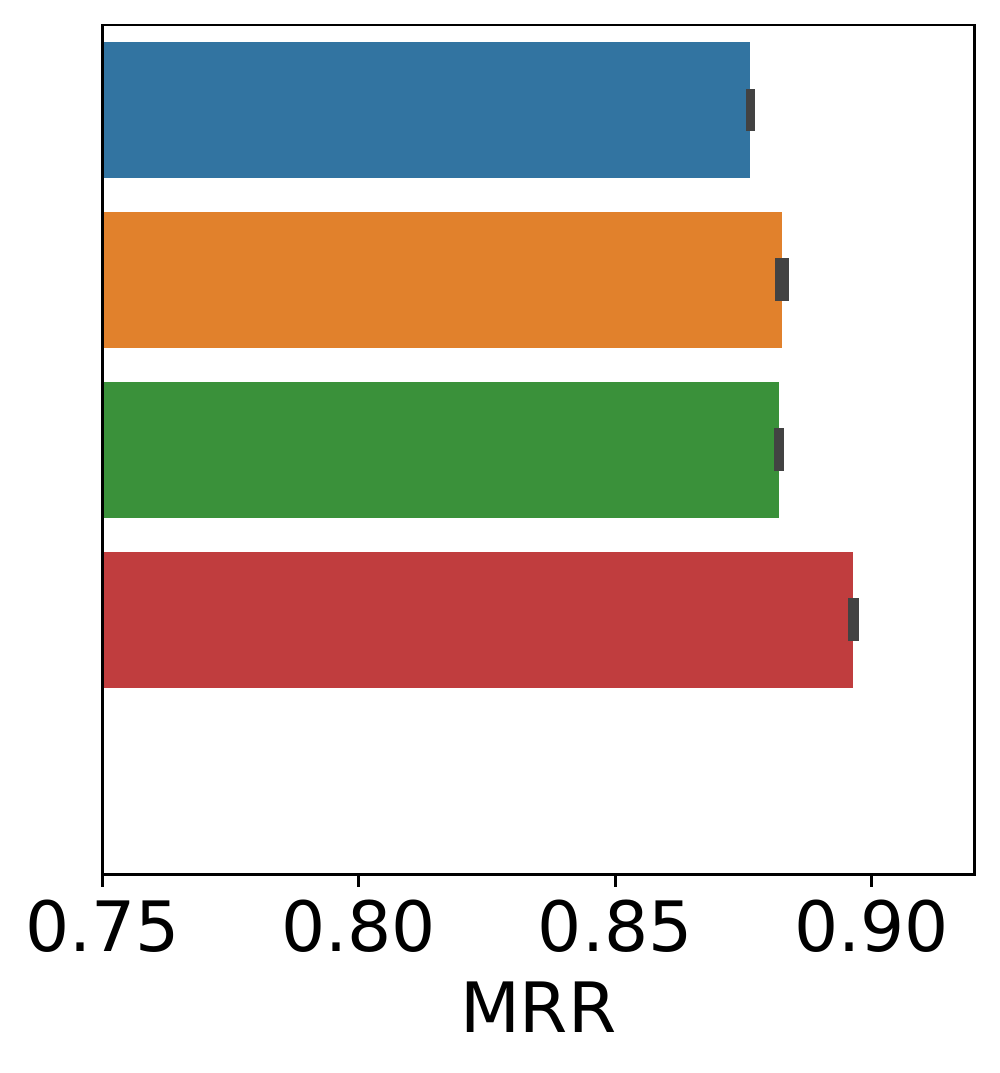}
         
\end{tabular}
\caption{A comparison of different mechanisms for processing AST structure in Transformer, in the anonymized setting. The numeric data for barplots is given in Table~\ref{tab:barplots_anon}. Please mind that the x-axis limits are different from Figure~\ref{fig:exp2a_full}.}
\label{fig:exp2a_anon}
\end{figure*}

The results are shown in Figure~\ref{fig:exp2a_anon}. The leading approaches are the same as in the full data setting: sequential relative attention in the VM and CC (value prediction) tasks, tree relative attention in the CC (type prediction) task, and almost similar performance of different mechanisms in the FN task.

\section{Tables and additional plots for comparing different structure-capturing mechanisms}
\label{app:numdata}
\begin{figure*}[t!]
\begin{tabular}{cccc}
         \hspace{0.2cm} Python: \hspace{0.55cm} Variable Misuse &  Function Naming &  Code Completion (values) & Code Completion (types)\\
          \includegraphics[height=0.15\linewidth]{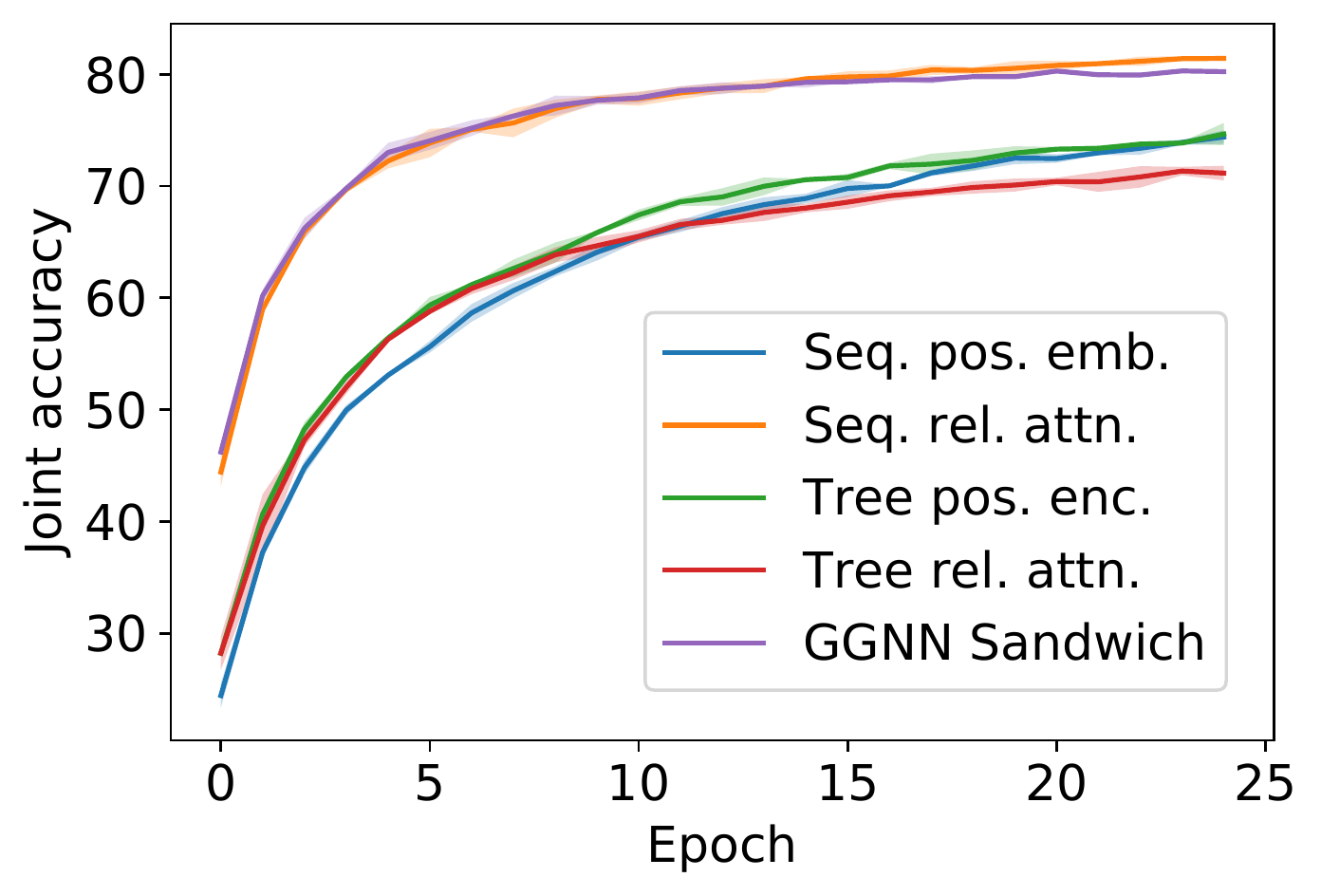} &
         \includegraphics[height=0.15\linewidth]{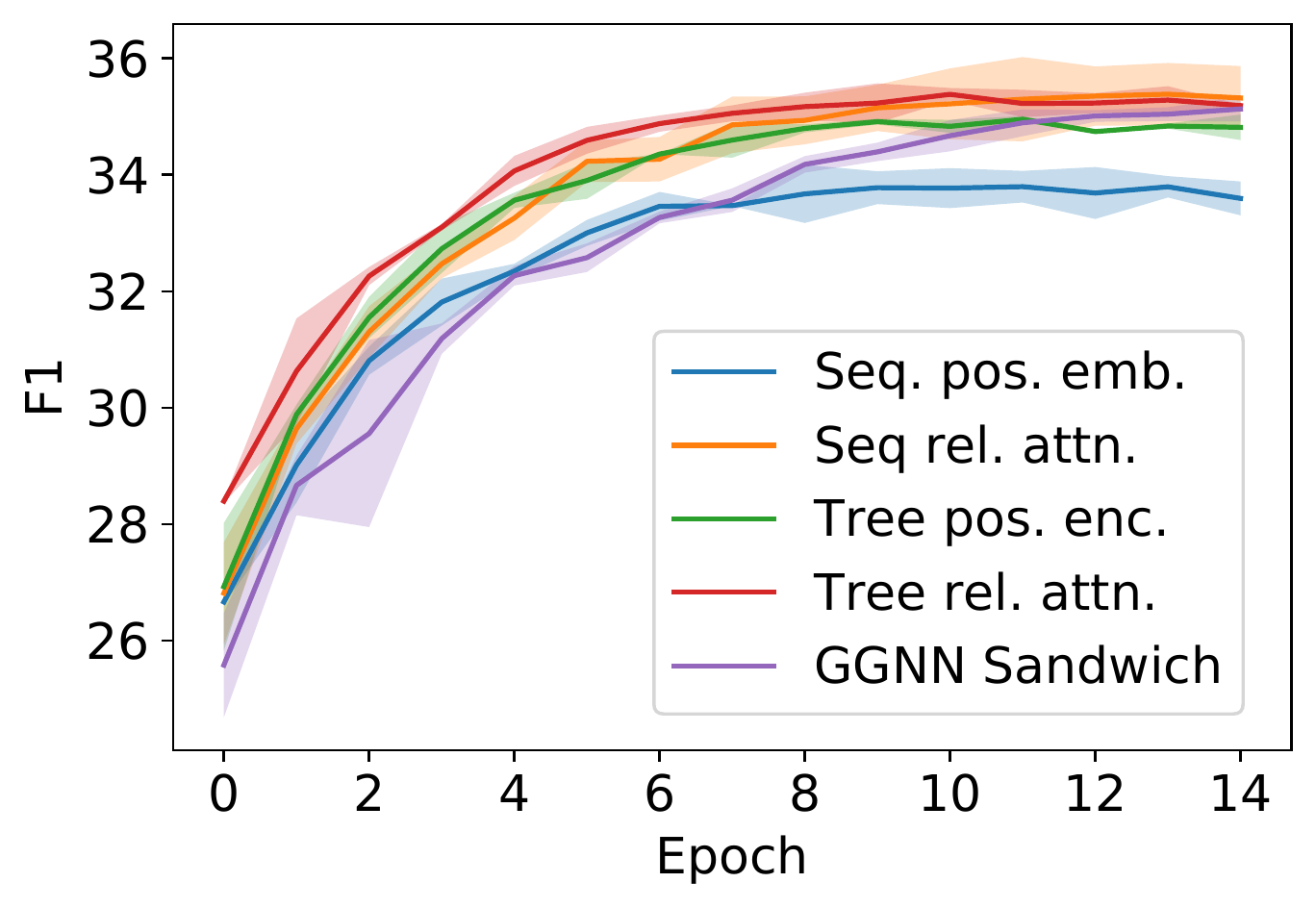} &
\includegraphics[height=0.15\linewidth]{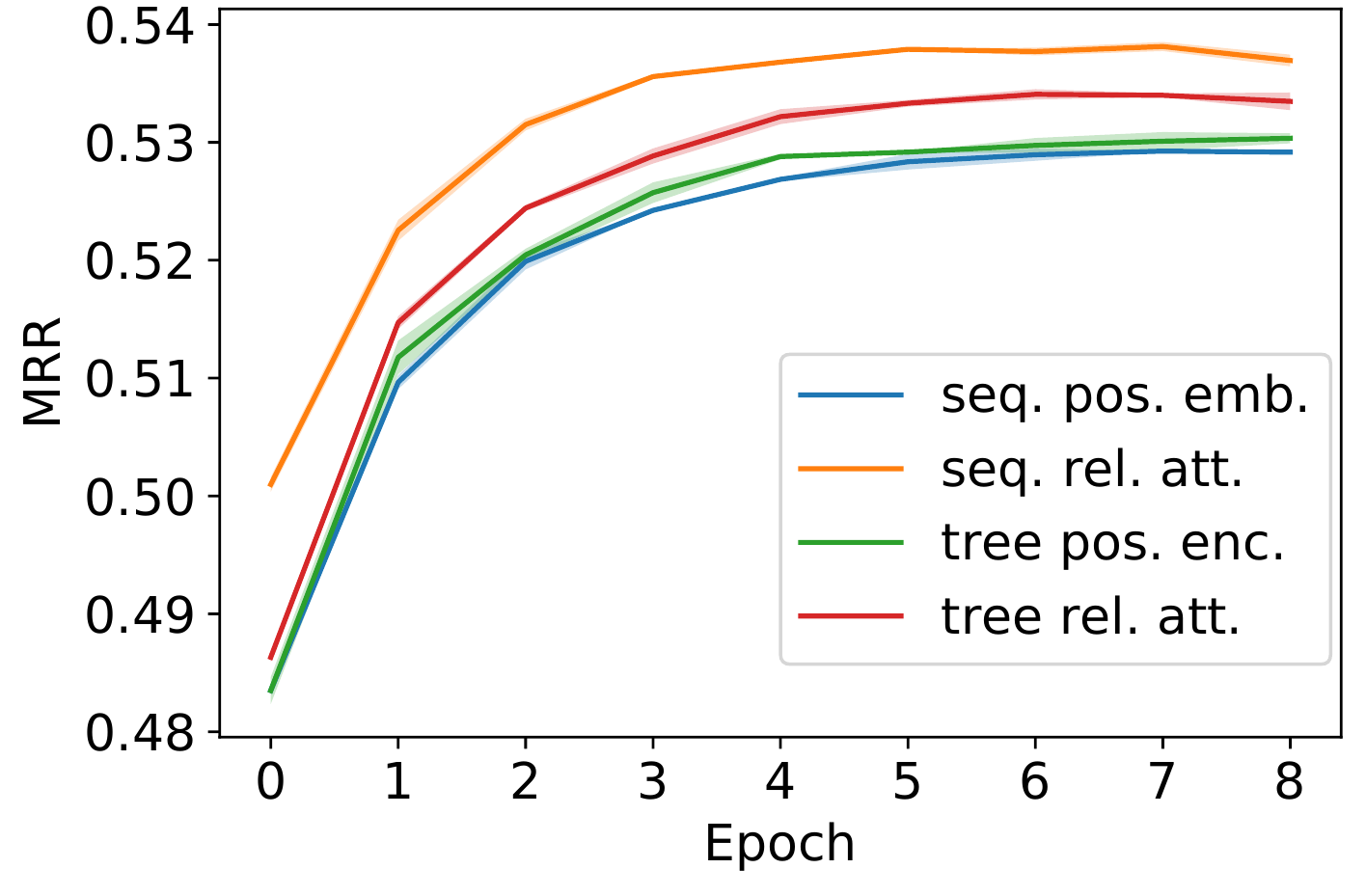} &
\includegraphics[height=0.15\linewidth]{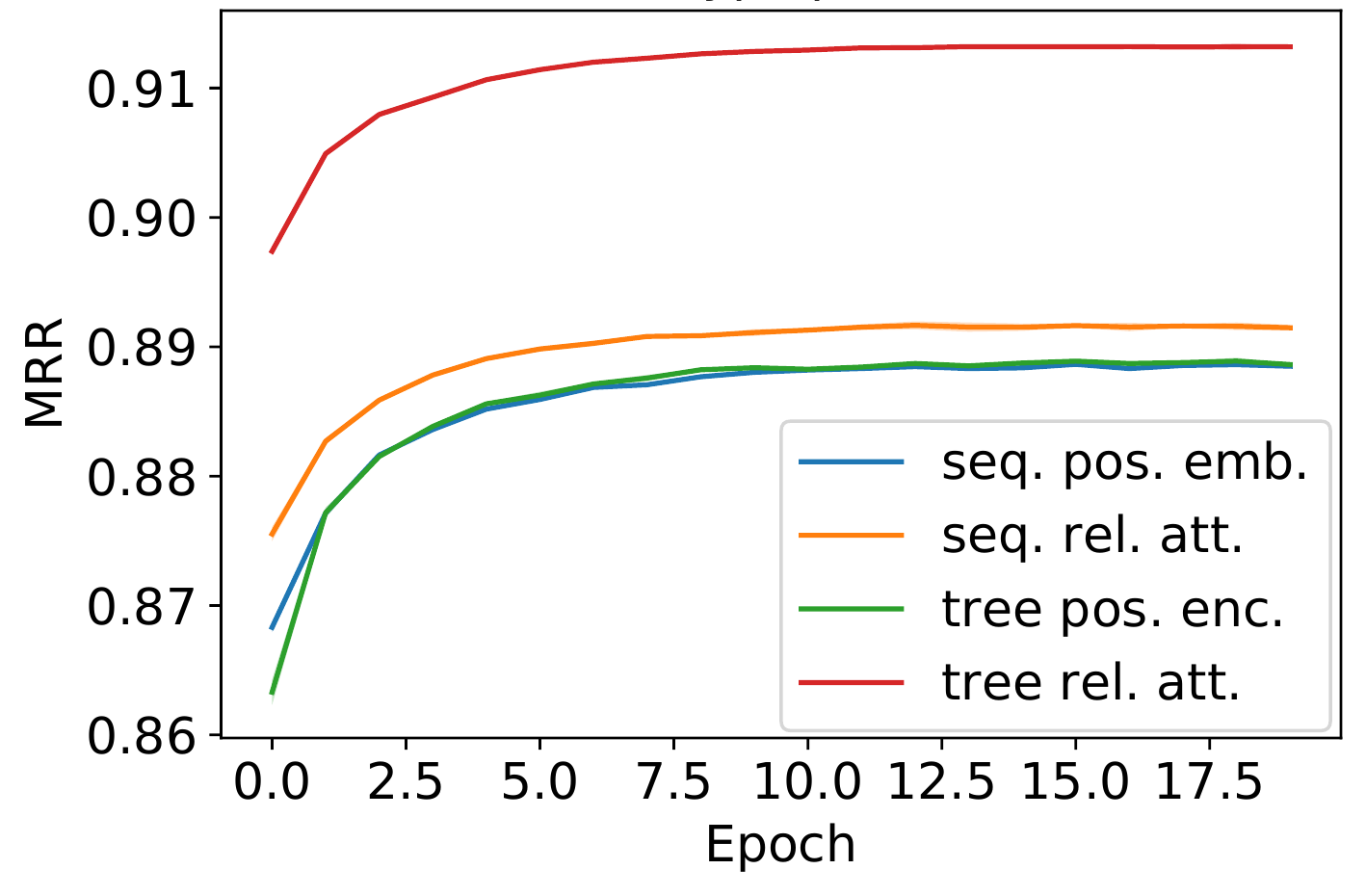}
\\
         JavaScript: \hspace{0.4cm} Variable Misuse \hspace{0.1cm} &  Function Naming  &  Code Completion (values)   &  Code Completion (types)  \\
         \includegraphics[height=0.15\linewidth]{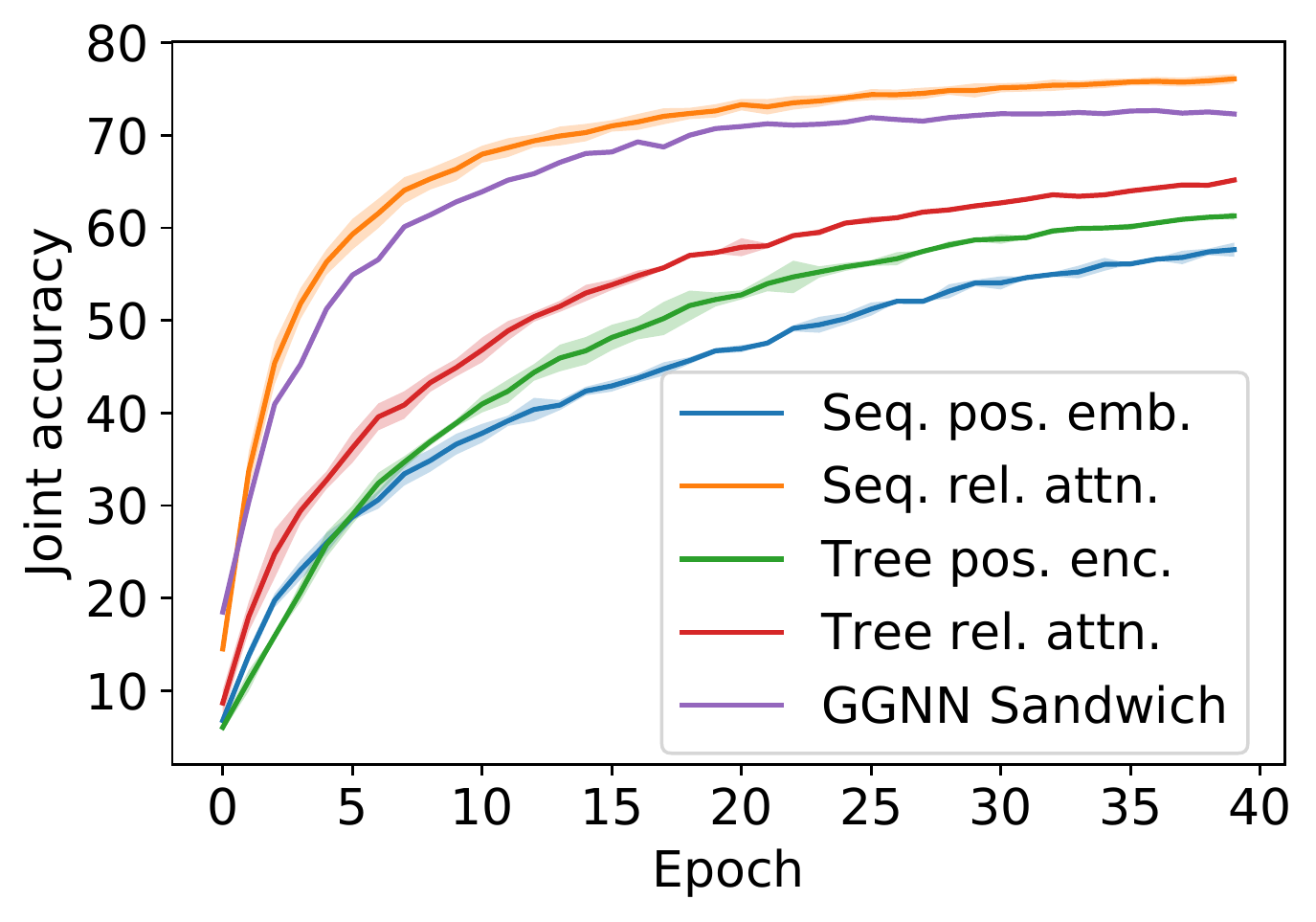} &
         \includegraphics[height=0.15\linewidth]{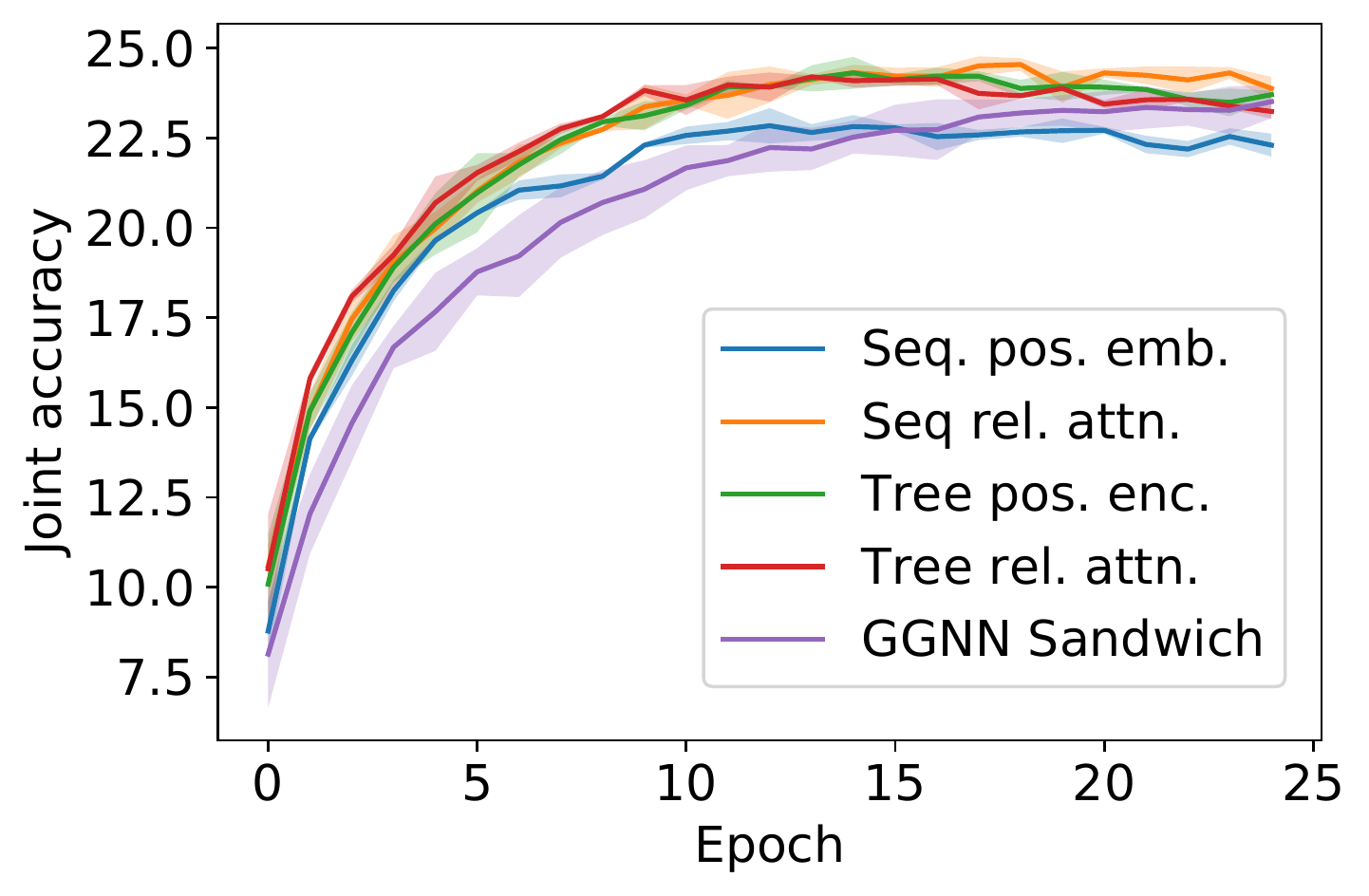} &
         \includegraphics[height=0.15\linewidth]{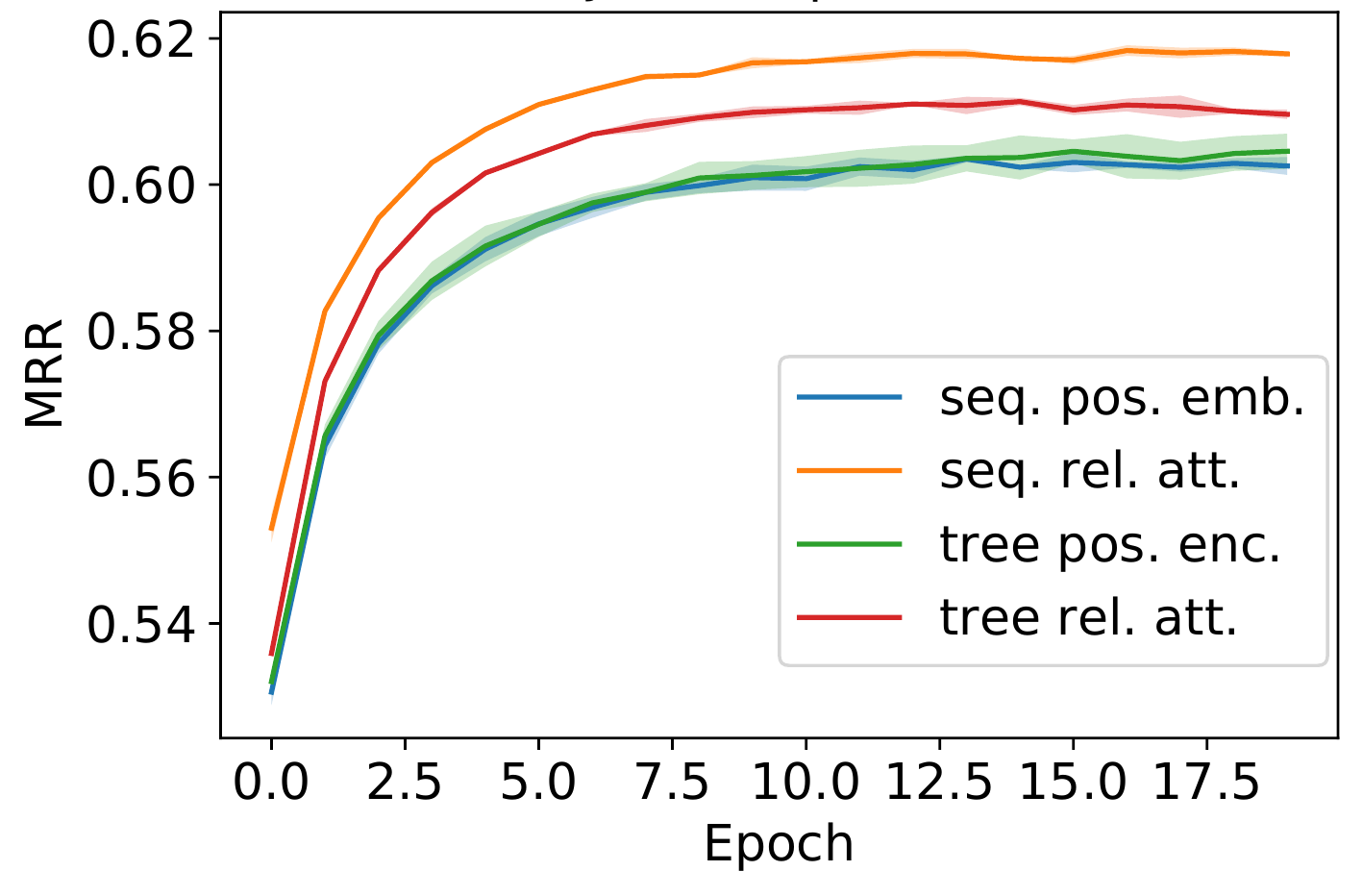} &
         \includegraphics[height=0.15\linewidth]{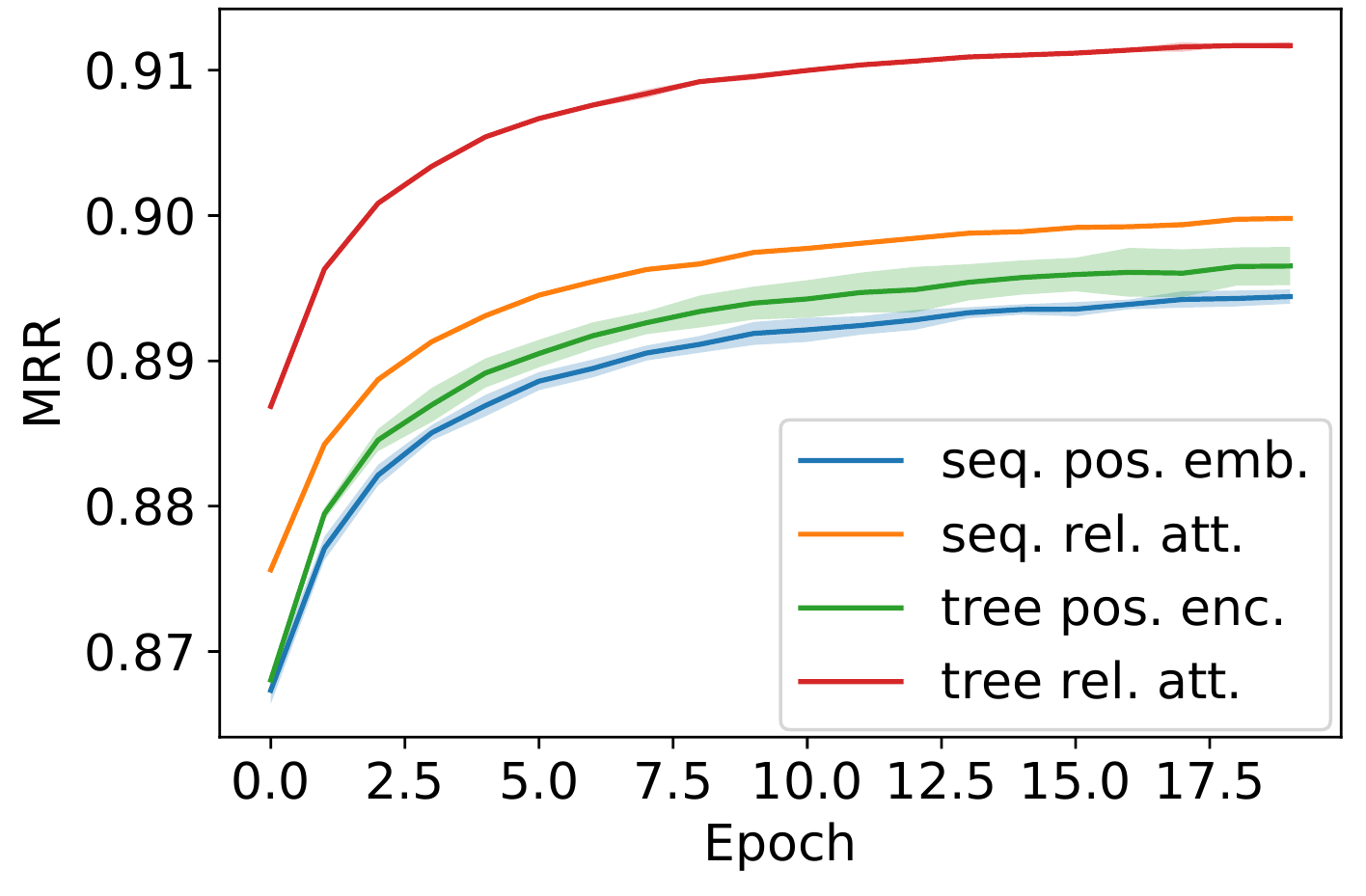}
         
\end{tabular}
\caption{A comparison of different mechanisms for processing AST structure in Transformer: test metrics by epochs.}
\label{fig:exp2_full_plots}
\end{figure*}

Table~\ref{tab:barplots_full} lists numerical data for barplots presented in Figure~\ref{fig:exp2a_full} in the main part of the paper (comparing AST-processing mechanisms in the full-data setting). Figure~\ref{fig:exp2_full_plots} visualizes the progress of test metrics during training for different AST-processing mechanisms in the full-data setting. This plots shows that the number of epochs we use is sufficient, i. e. the ordering of methods would not change and all models converged or almost converged. Table~\ref{tab:barplots_full} lists numerical data for barplots presented in Figure~\ref{fig:exp2a_anon} (comparing structure-capturing mechanisms in the anonymized setting).

\begin{table*}[h]
	\centering
	\caption{The numerical data for the comparison of different mechanisms for processing AST structure in Transformer, in the full-data setting.}
	\begin{tabular}{|c|cc|cc|cc|cc|}
	\hline
	&  \multicolumn{2}{c|}{\textbf{Variable misuse}} &  \multicolumn{2}{c|}{\textbf{Function naming}} &  \multicolumn{2}{c|}{\textbf{Code completion (Values)}} &  \multicolumn{2}{c|}{\textbf{Code completion (Types)}}\\  \midrule
	\textbf{Model} &  \textbf{PY} & \textbf{JS} &  \textbf{PY} & \textbf{JS} &  \textbf{PY} & \textbf{JS}&  \textbf{PY} & \textbf{JS}\\ \hline
	Seq. pos. emb. &  74.38$\pm$0.56 & 55.12$\pm$0.19 &  33.79$\pm$0.20 & 23.11$\pm$0.27 & 53.50$\pm$0.02 & 62.7$\pm$0.18      & 88.90$\pm$0.02 & 89.59$\pm$0.07 \\
	Seq. rel. attn.                 
	&  81.42$\pm$0.16 & 76.52$\pm$0.34 &  35.49$\pm$0.35 & 24.67$\pm$0.18 & 54.53$\pm$0.07 & 64.11$\pm$0.03       & 89.22$\pm$0.02 & 90.14$\pm$0.002\\
	Tree pos. enc.                 
	&  74.65$\pm$0.99 & 61.29$\pm$0.41 &  34.95$\pm$0.40 & 24.45$\pm$0.21 & 53.82$\pm$0.007 & 62.70$\pm$0.45     &  88.96$\pm$0.005 & 89.71$\pm$0.21\\
	Tree rel. attn.                 
	&  71.33$\pm$0.54 & 65.17$\pm$0.09 &  35.19$\pm$0.17 & 24.24$\pm$0.11 & 54.00$\pm$0.10 & 63.39$\pm$0.11  & 91.36$\pm$0.01 & 91.2$\pm$0.03 \\
	GGNN Sandwich                  
	&  80.23$\pm$0.15 & 72.58$\pm$0.14 &  35.40$\pm$0.21 & 23.57$\pm$0.50 & N/A & N/A & N/A& N/A \\ \hline
	\end{tabular}
	\label{tab:barplots_full}
\end{table*}

\begin{table*}[h]
	\centering
	\caption{The numerical data for the comparison of different mechanisms for processing AST structure in Transformer, in the anonymized setting.}
	\begin{tabular}{|c|cc|cc|cc|cc|}
	\hline
	&  \multicolumn{2}{c|}{\textbf{Variable misuse}} &  \multicolumn{2}{c|}{\textbf{Function naming}} &  \multicolumn{2}{c|}{\textbf{Code completion (Values)}} &  \multicolumn{2}{c|}{\textbf{Code completion (Types)}}\\  \midrule
	\textbf{Model} &  \textbf{PY} & \textbf{JS} &  \textbf{PY} & \textbf{JS} &  \textbf{PY} & \textbf{JS} &  \textbf{PY} & \textbf{JS} \\ \hline
	Seq. pos. emb.
	&  77.21$\pm$0.02 & 66.19$\pm$0.65 &  23.40$\pm$0.51 & 12.24$\pm$0.12       & 58.18$\pm$0.08   & 59.71$\pm$0.01        &  88.10$\pm$0.04 &  87.64$\pm$0.01 \\
	Seq. rel. attn.                 
	&  81.83$\pm$0.12 & 78.53$\pm$0.17 &  24.79$\pm$0.15 & 13.70$\pm$0.19       & 58.78$\pm$0.05   & 60.83$\pm$0.08        &  88.44$\pm$0.03  & 88.25$\pm$0.06 \\
	Tree pos. enc.                 
	&  74.86$\pm$0.19 & 61.07$\pm$1.82 &  24.52$\pm$0.25 & 13.43$\pm$0.05       & 58.27$\pm$0.09   & 60.31$\pm$0.01        &  88.23$\pm$0.05  & 88.19$\pm$0.01 \\
	Tree rel. attn.                 
	&  67.35$\pm$0.03 & 65.47$\pm$0.44 &  23.94$\pm$0.20 & 12.91$\pm$0.12       & 58.45$\pm$0.16   & 60.45$\pm$0.05        &  90.72$\pm$0.04  & 89.64$\pm$0.03 \\
	GGNN Sandwich                
	&  77.22$\pm$0.08 & 69.64$\pm$0.11 &  24.91$\pm$0.11 & 13.11$\pm$0.22 & N/A & N/A & N/A& N/A \\ \hline
	\end{tabular}
	\label{tab:barplots_anon}
\end{table*}

\clearpage
\section{Attention maps for different structure-capturing mechanisms in Transformer}
\label{app:attention_maps}

\begin{figure*}[t!]
\centering

 \centering
 \includegraphics[width=0.95\linewidth]{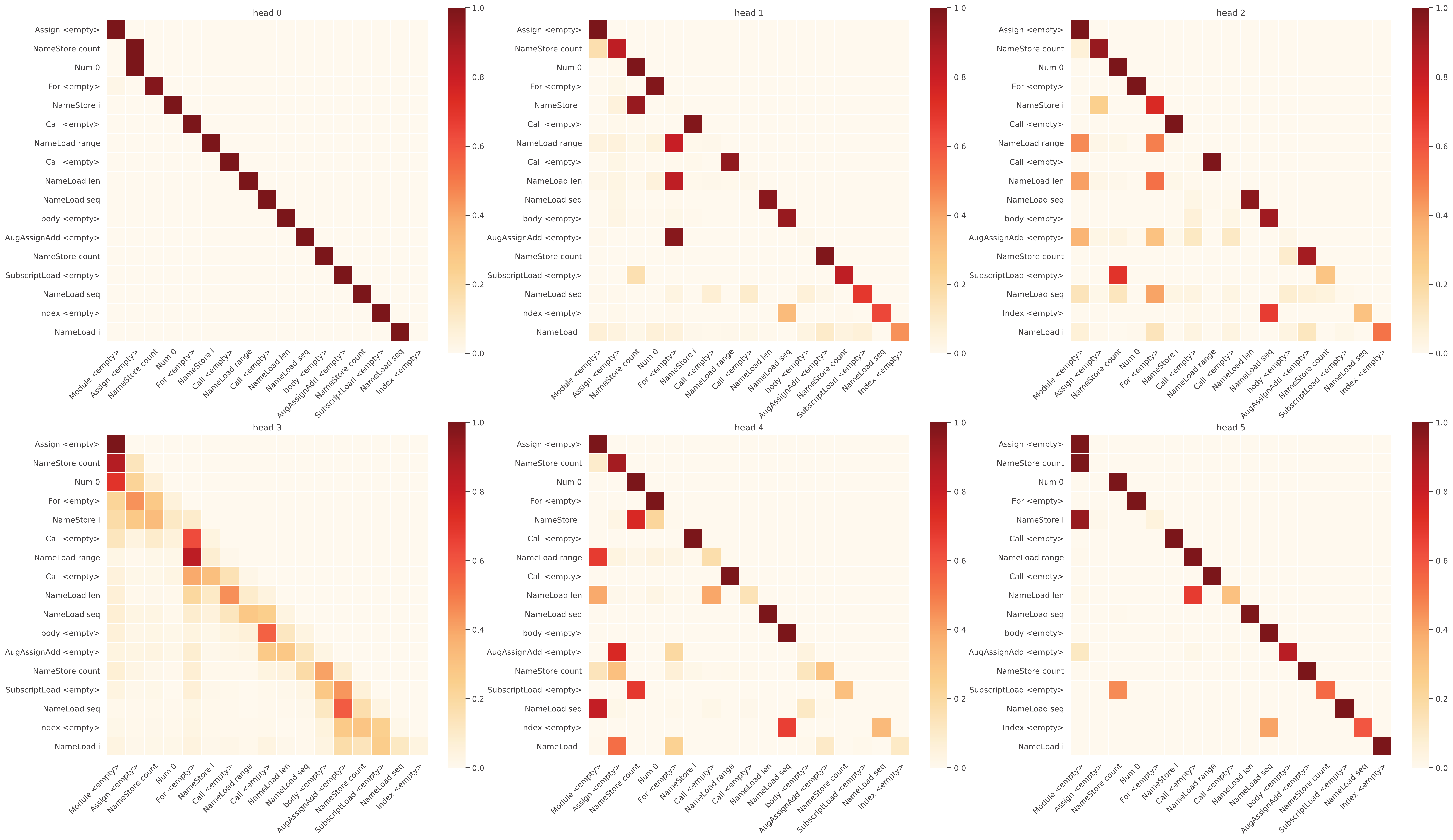}
 \caption{Attention maps for the 1st layer  of \textit{Syntax} model, sequential positional embeddings, code completion task. Y axis: what the model predicts, X axis: which token the model attends to.}
 \label{fig:pos_emb}

\hfill

 \centering
 \includegraphics[width=0.95\linewidth]{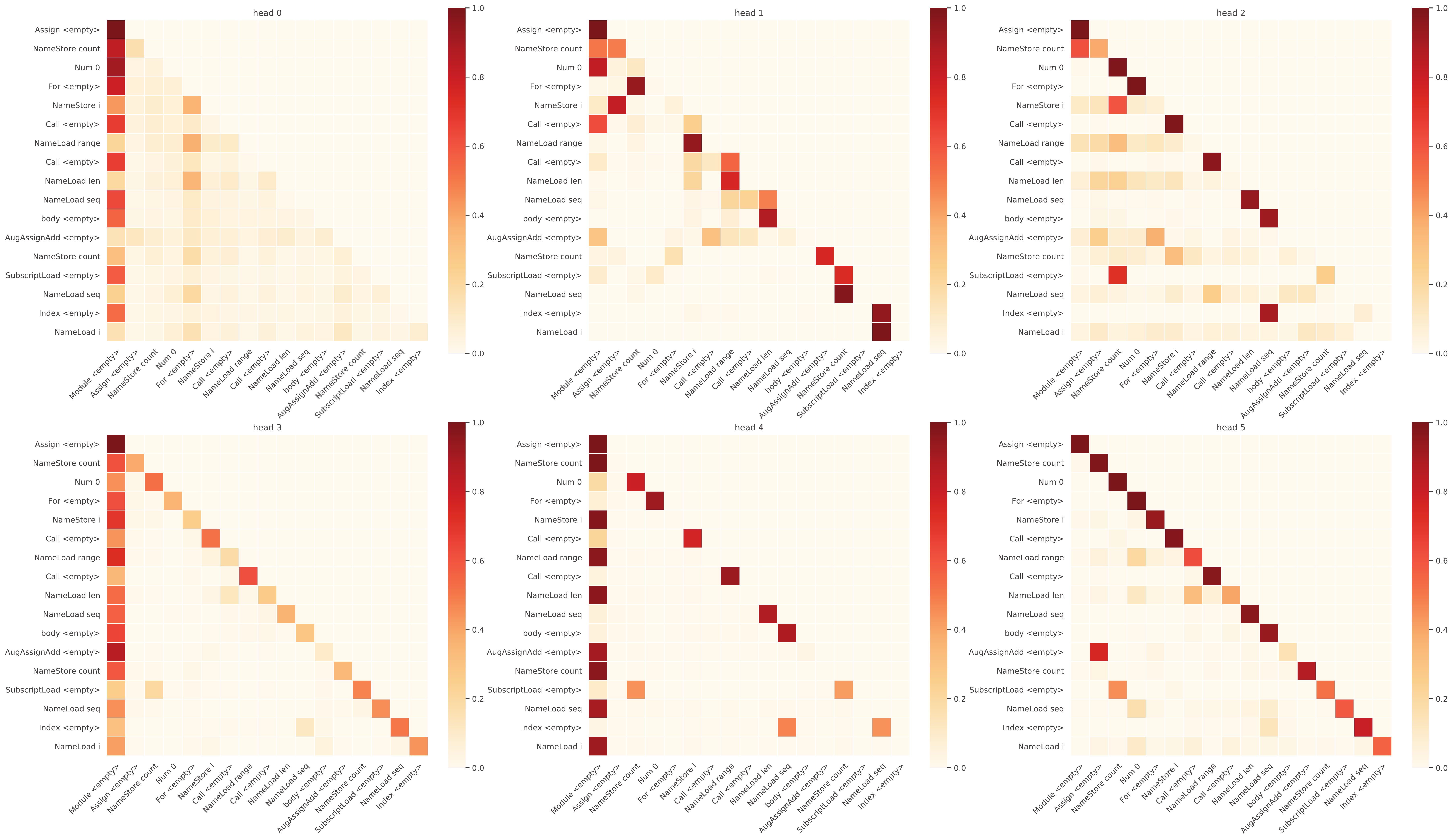}
 \caption{Attention maps for the 1st layer of \textit{Syntax} model, tree positional encodings, code completion task.
  Y axis: what the model predicts, X axis: which token the model attends to.}
 \label{fig:tree_enc}

\end{figure*}

\begin{figure*}[t!]
\centering

 \centering
 \includegraphics[width=0.95\linewidth]{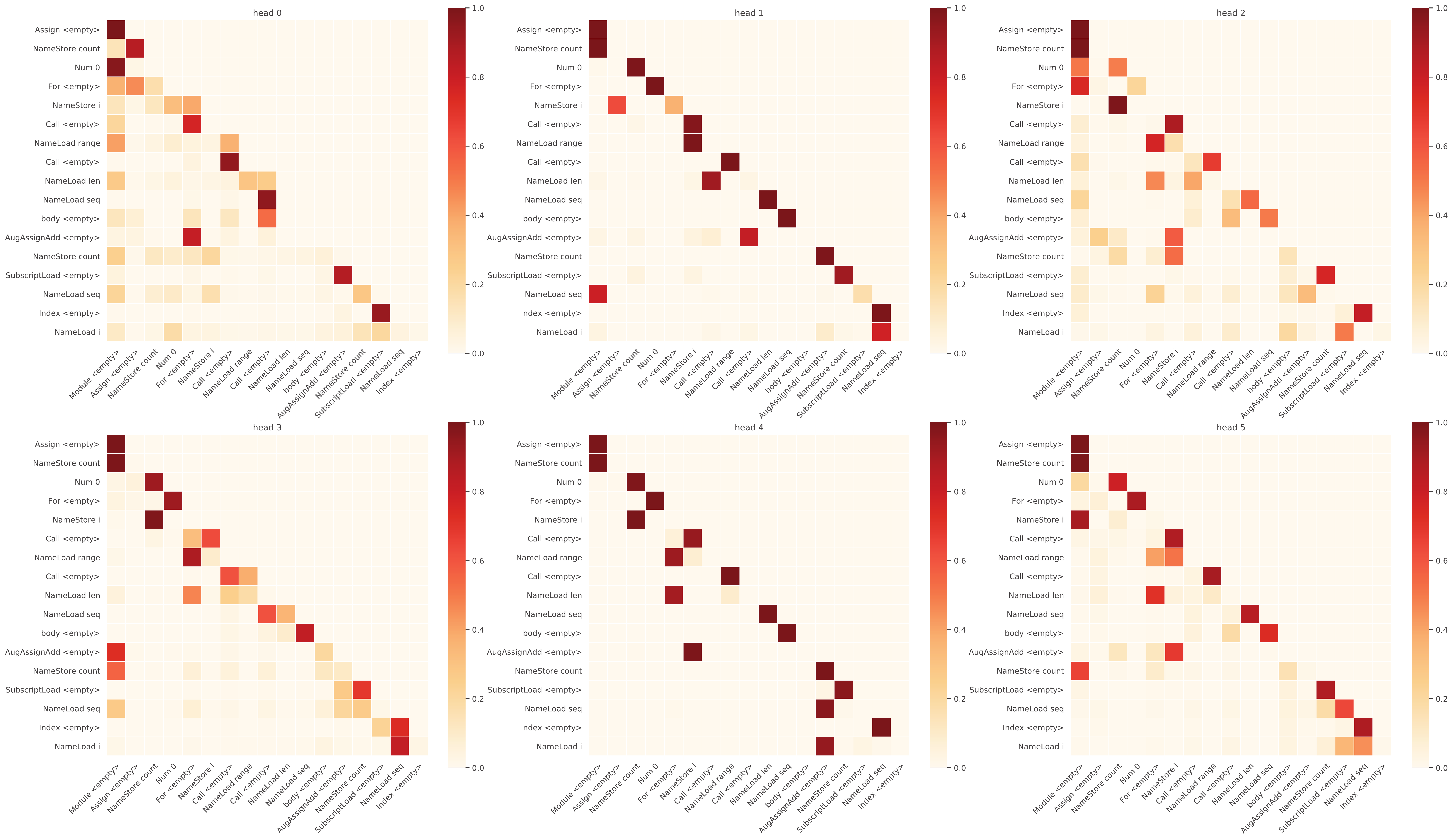}
 \caption{Attention maps for the 1st layer of \textit{Syntax} model, sequential relative attention, code completion task.
  Y axis: what the model predicts, X axis: which token the model attends to.}
 \label{fig:seq_rel}

 \centering
 \includegraphics[width=0.95\linewidth]{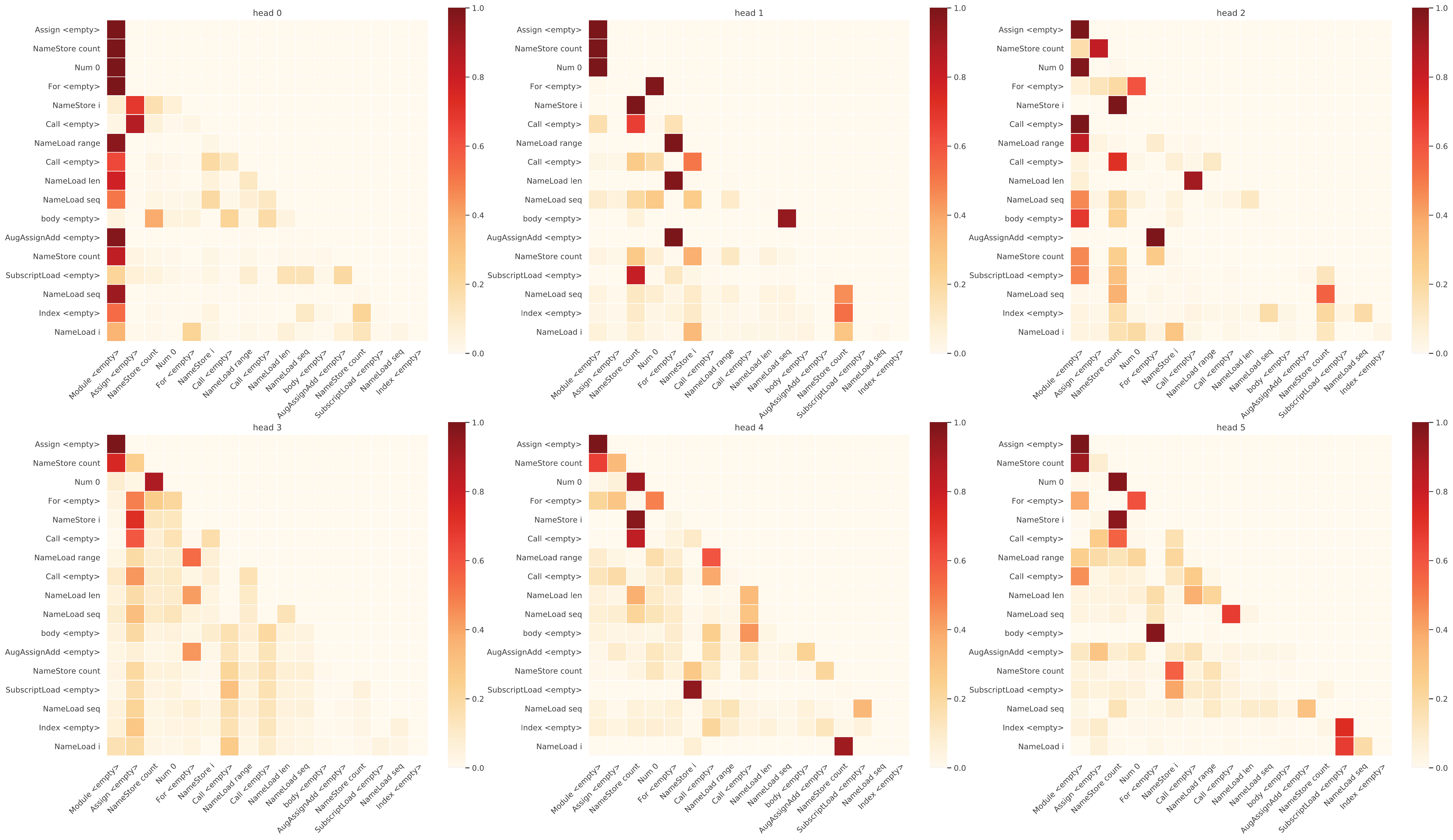}
 \caption{Attention maps for the 1st layer of \textit{Syntax} model, tree relative attention, code completion task.
  Y axis: what the model predicts, X axis: which token the model attends to.}
 \label{fig:tree_rel}

\end{figure*}

In Figures ~\ref{fig:pos_emb}, \ref{fig:tree_enc}, \ref{fig:seq_rel}, \ref{fig:tree_rel}  we present attention maps for different structure-capturing mechanisms on the code completion task in the anonymized setting. We visualize attention maps for the first layer since low level interactions should reflect the differences in considered mechanisms. We observe that all mechanisms except tree relative attention mostly attend to the last predicted tokens. Transformer with tree positional encodings (figure~\ref{fig:tree_enc}) 
always pays significant attention to the root node, that is easy to find because of zero tree encoding, or to some ``anchor'' nodes, e. g. \verb|For| node in the first map of Figure~\ref{fig:tree_enc}. In other words, tree positional encodings allow emphasizing important nodes in the tree. Also this model is able to distinguish children numbers.
Transformer with tree relative attention (figure~\ref{fig:tree_rel}) often watches at siblings of the node to be predicted, but this model cannot differentiate child numbers, by construction. The attention maps for this model are smoother than for other models, because this model introduces multiplicative coefficients to the weights after softmax in self attention.


\begin{figure}
\begin{tabular}{c}
\includegraphics[scale=1]{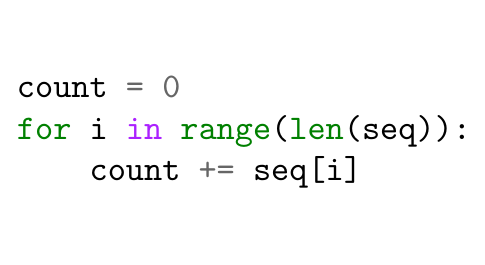} \\
\includegraphics[width=0.95\linewidth]{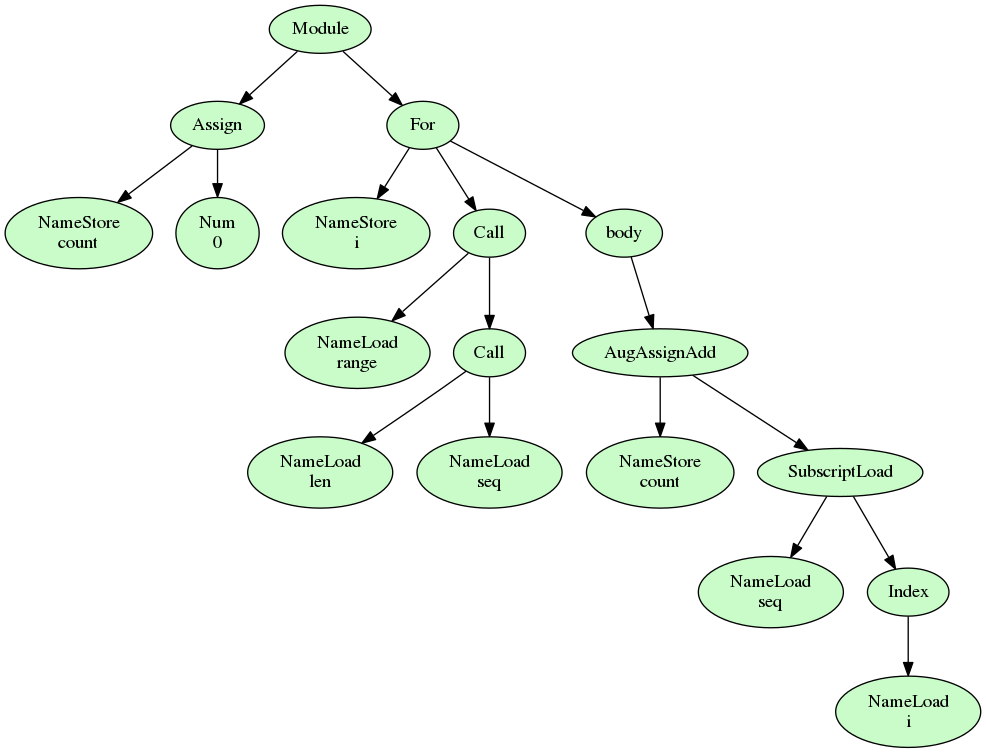}
\end{tabular}
\caption{Code snippet (and its AST) used to visualize attention maps.}
\label{fig:snippet}
\end{figure}

\end{document}